\documentclass[11pt]{article}
\usepackage{vqil}

\usepackage{amsmath,amsfonts,bm}

\def\eqref#1{equation~\ref{#1}}

\def\1{\bm{1}}

\DeclareMathAlphabet{\mathsfit}{\encodingdefault}{\sfdefault}{m}{sl}
\SetMathAlphabet{\mathsfit}{bold}{\encodingdefault}{\sfdefault}{bx}{n}

\usepackage{graphicx}
\usepackage{xcolor}
\usepackage{makecell}
\usepackage{booktabs}
\usepackage{amsmath,amssymb}
\usepackage{multirow}
\usepackage{float}
\usepackage{algorithm}
\usepackage{algorithmic}
\usepackage{enumitem}

\title{VQIL: Visible Quantum Interference Learning\\ for Perturbation Response Ranking on Graphs}

\shorttitle{VQIL: Visible Quantum Interference Learning}

\date{}

\hypersetup{pdftitle={VQIL: Visible Quantum Interference Learning for Perturbation Response Ranking on Graphs},pdfauthor={Seokjun Lee, Sergio A. Ortega, Daniel K. Park, Namshik Han}}

\author{%
Seokjun Lee$^{1,2,3}$, Sergio A. Ortega$^{4}$, Daniel K. Park$^{4,5,6}$, Namshik Han$^{1,2,3,6,7,8,9,10,11,12,*}$
\begin{affiliations}
\affiliation{1}{Milner Therapeutics Institute, University of Cambridge, Cambridge, United Kingdom}
\affiliation{2}{Cambridge Stem Cell Institute, University of Cambridge, Cambridge, United Kingdom}
\affiliation{3}{Department of Applied Mathematics and Theoretical Physics, Cambridge Centre for AI in Medicine, University of Cambridge, Cambridge, United Kingdom}
\affiliation{4}{Department of Statistics and Data Science, Yonsei University, Seoul, Republic of Korea}
\affiliation{5}{Department of Applied Statistics, Yonsei University, Seoul, Republic of Korea}
\affiliation{6}{Department of Quantum Information, Yonsei University, Seoul, Republic of Korea}
\affiliation{7}{Urological Malignancies Virtual Institute, University of Cambridge, Cambridge, United Kingdom}
\affiliation{8}{Department of Nano Biomedical Engineering, Advanced Science Institute, Yonsei University, Seoul, Republic of Korea}
\affiliation{9}{Institute for Basic Science (IBS) Center for Nanomedicine, Yonsei University, Seoul, Republic of Korea}
\affiliation{10}{Department of Artificial Intelligence, Yonsei University, Seoul, Republic of Korea}
\affiliation{11}{College of Pharmacy, Yonsei University, Incheon, Republic of Korea}
\affiliation{12}{Institute for Convergence Research and Education in Advanced Technology, Yonsei University, Seoul, Republic of Korea}
\par\vskip 0.6em
\noindent$^{*}$Corresponding author. Email: \href{mailto:nh417@cam.ac.uk}{nh417@cam.ac.uk}
\end{affiliations}}

\begin{document}

\maketitle

\begin{abstract}
Learning how local perturbations affect a network is important for identifying which components drive or absorb a system's response. We formulate this as seed-conditioned response ranking: given a perturbation at one seed node, ranking all other nodes by response strength. This is challenging when the model is given only the interaction topology, while the mechanism governing how perturbations propagate must be learned from data. We present VQIL (Visible Quantum Interference Learning), a trainable Szegedy quantum walk for seed-conditioned ranking. Edge weights, link phases, and per-node phase rotations are optimized end-to-end by backpropagation while the walk operator remains unitary throughout training. Node scores are obtained from a geometrically discounted coherent sum of quantum-walk amplitudes, enabling quantum interference before the final squared-norm readout. Because each learned parameter is attached to a named edge or node, the trained model provides directly inspectable hypotheses about the roles of individual graph components in the propagation mechanism. On AC power-flow perturbations over six transmission grids, with electrical parameters withheld from the model, VQIL improves Spearman rank correlation from $0.147$ for the untrained walk to $0.831$ and outperforms an identically trained random-walk model. On gene perturbation over five protein--protein interaction networks with Perturb-seq supervision, VQIL improves rank correlation from $0.171$ to $0.620$, outperforming the strongest learned classical and neural baselines. These results show that trainable quantum-walk interference provides a scalable and interpretable inductive bias for learning seed-conditioned response rankings on graphs.

\end{abstract}

\section{Introduction}
\label{sec:intro}

Graphs provide a natural abstraction for representing complex systems, where many foundational questions reduce to seed-conditioned ranking: if one node of the system is perturbed, how strongly do the other nodes respond, and through what mechanism? Given a graph whose edges encode known interactions and a perturbed seed node, the task is to rank all remaining nodes by their downstream response magnitude. Instances range from in silico gene perturbation---ranking transcriptomic shifts when a target gene is knocked down---to infrastructure resilience analysis, such as identifying which buses in a power grid re-equilibrate under localized load injection. The task is substantive because the observed topology typically records only structural connectivity, not transmission capacity or functional dynamics. Two things are consequently asked of a model. It must learn the underlying transmission mechanism from perturbation-response supervision, and keep this mechanism exposed for scientific interpretation.

\medskip
Graph neural networks (GNNs) \citep{kipf2017,velickovic2018} are the dominant learning framework for graph-structured data and have ample capacity to learn complex perturbation mechanisms. However, they struggle to expose them: the seed typically enters through node attributes, and training optimizes shared transformation operators across the topology. Consequently, the learned mechanism remains distributed across continuous latent representations rather than expressed as explicit, interpretable interaction rules. For seed-conditioned ranking, however, a simpler family of methods remains highly relevant: \emph{graph propagation}. Propagation methods spread unit mass from the seed along edges and score nodes by the mass they accumulate. Random walk with restart is the canonical example and underpins network-propagation methods across several domains \citep{page1999,tong2006}. Supervised extensions learn the walk itself from training queries, via per-edge conductances, feature-based edge strengths, or per-node restart probabilities \citep{agarwal2006, backstrom2011,jin2019}, preserving a particularly useful correspondence between model and graph: the seed defines the initial condition, propagation occurs directly on the observed topology, and learned quantities modulate named edges or nodes. What is learned thus lives on the graph and can be read as a hypothesis about which components most strongly transmit the perturbation. This follows the philosophy of visible neural networks \citep{ma2018,yu2018}, where interpretability is achieved by construction rather than by post hoc explanation \citep{rudin2019}. We push this correspondence to its limit by enforcing strict architectural transparency:

\newtheorem{definition}{Definition}
\begin{definition}[Visibility]
\label{def:visibility}
A graph-learning model is \emph{visible} if every learned parameter is a scalar attached to a named edge or node of the input graph, so that each parameter can be interpreted as a hypothesis about the role of that component in the underlying mechanism.
\end{definition}

The propagation dynamics themselves, however, remain a limitation. Accurately ranking perturbation responses often requires combining the contributions of many parallel regulatory paths, which a classical walk can only accumulate as non-negative probability mass. Quantum walks provide a natural extension of this propagation view. Unlike classical diffusion, where path contributions are combined monotonically, quantum walks propagate complex amplitudes that combine coherently before probabilities are formed, allowing constructive and destructive interference \citep{aharonov2001,szegedy2004}. This provides a richer hypothesis class for modeling how perturbations propagate through a graph. By making the graph-phased Szegedy quantum walk trainable \citep{ortega2025}, we turn quantum interference into a learnable propagation mechanism:

\begin{definition}[Quantum interference learning]
\label{def:qil}
Quantum interference learning is supervised learning over the hypothesis space of unitary walks on a given graph: gradient descent on the walk's own edge and node parameters selects the interference pattern whose readout best explains the observed responses.
\end{definition}

We introduce \emph{visible quantum interference learning} (VQIL), a trainable Szegedy quantum walk for seed-conditioned response ranking. VQIL learns edge weights, edge phases, and node phases end-to-end from supervision while preserving unitarity throughout training. The resulting model combines interference-based propagation with parameter-level visibility: the walk is optimized to rank perturbation responses accurately, while its learned quantities remain attached to named graph components and can therefore be examined mechanistically. Our contributions are threefold:

\begin{itemize}\itemsep2pt
\item \textbf{A trainable Szegedy quantum walk.} We make the phased walk operator---per-edge weights, per-edge link phases, per-node phase rotations---fully differentiable and learnable. The standard walk provides the initialization, unitarity is preserved at every point of training, and the resulting implementation scales at $O(m)$ cost per step for $m$ directed edges.

\item \textbf{A coherent-resolvent readout.} Node scores are derived from a geometrically discounted sum of walk amplitudes, preserving rich quantum interference prior to the final squared-norm measurement. The discount replaces a hard temporal cutoff with a smooth, convergent horizon, yielding a resolvent of the walk operator and a differentiable objective across all walk parameters.

\item \textbf{Visible mechanistic interpretation.} Every learned quantity is a scalar on a named edge or node of the input graph. The trained model therefore constitutes an explicit, graph-level hypothesis about the propagation mechanism.  We show that the trained fields carry information about withheld properties of the underlying systems, turning the model into a hypothesis-generating tool for domain experts.
\end{itemize}

\section{Problem statement}
\label{sec:problem}

Let $G = (V, E)$ be an undirected, unweighted, connected simple graph with $n=|V|$ nodes, and let
$m$ denote the number of directed edges (where each undirected edge is represented in both directions). A \emph{query} is a pair
$(s, y_s)$, where $s \in V$ is the seed node at which the system is perturbed, and the label $y_s \in \mathbb{R}^{|V|-1}_{\ge 0}$ contains the absolute response magnitude measured at every other node.
A model is a parameterized mapping $f_\Theta : (G, s) \mapsto x \in \mathbb{R}^{|V|}_{\ge 0}$, where $\Theta$ denotes the full parameter set. The model assigns a response score to every node conditioned on the perturbed seed. Predictive performance is evaluated by how well the rank ordering of the predicted scores $x$ (excluding the seed) aligns with the ordering of
the ground-truth measured responses $y_s$. Training minimizes a differentiable ranking loss over a set of training queries, while evaluation uses only queries whose seeds were held out during training.

\medskip
The defining property of the task is that \emph{the supervision carries functional dynamics that the observed
graph does not natively encode}. The true response $y_s$ is generated by an external physical or biological mechanism, whereas the input graph records only structural connectivity. Consequently, the graph defines the model's \emph{hypothesis space}, not its answer sheet. Any predictive accuracy above a naive topological baseline must come from the learned mechanism. Because the model is strictly visible, whatever mechanism is learned is directly readable from $\Theta$ as a quantifiable statement about the underlying system.

\section{Related work}
\label{sec:related}

\textbf{Trainable propagation and quantum walks.} Random-walk propagation is the standard tool for
seed-conditioned prioritization: random walk with restart \citep{tong2006, page1999}, which underlies network propagation analyses in biology and other domains \citep{cowen2017, han2021}, diffuses scalar probability mass. Supervised extensions learn this diffusion from training queries \citep{agarwal2006,backstrom2011,jin2019}. Quantum walks substitute this classical diffusion with unitary amplitude evolution and have been applied as fixed topological propagators for node centrality \citep{paparo2012,paparo2013} and disease-network analysis \citep{saarinen2024, park2026longcovid}. While the theoretical capacity for arbitrary per-node and link phases in Szegedy walks is described \citep{ortega2023,ortega2025}, alongside scalable simulators \citep{ortega2024squwals,ortega2026sqwlib}, existing trainable quantum walk neural networks learn coin operators that act on continuous node features \citep{dernbach2018,dernbach2019}. VQIL instead optimizes the walk's native per-edge and per-node parameters directly and extracts scores through a differentiable coherent resolvent.

\medskip
\textbf{Message-passing and geometric graph models.} Standard message-passing GNNs \citep{kipf2017, velickovic2018} optimize shared filters acting on continuous node features. Consequently, perturbation mechanisms are embedded in feature-space transformations rather than in explicit graph operators. Our approach aligns more closely with geometric graph learning, specifically with sheaf neural networks \citep{bodnar2022}, which learn per-edge restriction maps to modulate message passing; their $O(2)$ variant, which learns planar rotations, is the real-valued, feature-space analog of VQIL's complex-valued $U(1)$ link phases. To isolate the capacity of the propagation mechanism itselft, we evaluate against both families without external node attributes, so that all models rely only on the topology.

\medskip
\textbf{Structurally visible architectures.} To keep mechanistic readouts faithful to the model, VQIL avoids post hoc explanation methods \citep{rudin2019} in favor of intrinsically interpretable architecture. Visible neural networks achieve this by structurally constraining neural connectivity to mirror known biological ontologies or hierarchies \citep{ma2018,yu2018}. VQIL instantiates this visibility constraint strictly at the graph level: by enforcing a bijective mapping between the learnable parameters and the physical network's nodes and edges, the optimized walk parameters act as direct hypotheses (e.g., transmission strengths or node-level response propensities) that can be audited independently.

\section{Methods}
\label{sec:methods}

\subsection{Learned parameters and unitary propagation}
\label{sec:walk}

VQIL is built on the graph-phased formulation of the Szegedy quantum walk \citep{ortega2025}. The walker's state lives on the space spanned by the $m$ directed edges of the graph: register 1 holds the node the walker occupies, register 2 the
node it points at. One step of the walk is $U = SR$, where $S$ swaps the two registers, and
\begin{equation}
	R \;=\; \sum_i \big(1 - e^{i\theta_i}\big)\, |\psi_i\rangle\langle\psi_i| \;-\; I, \quad |\psi_i\rangle \;=\; |i\rangle_1 \otimes \sum_{k} e^{i \varphi_{ik}} \sqrt{p_{ki}}\, |k\rangle_2.
	\label{eq:walkop}
\end{equation}
From this operator we define the following trainable parameters of the walk.

\medskip
\textbf{Edge weights.} The weight $p_{ji}$ is the probability of jumping from node $i$ to node $j$. Each undirected edge $\{i,j\}$ carries a single, symmetric log-weight $\omega_{ij} = \omega_{ji} \in [-2, 2] $. The column-stochastic transition matrix is the softmax normalization over each node's incident edges,
\begin{equation}
p_{ji} \;=\; \frac{\exp(\omega_{ij})}{\sum_{k :\, \{i,k\} \in E} \exp(\omega_{ik})},
\label{eq:weights}
\end{equation}
so columns sum to one for every parameter value, and $\omega \equiv 0$ recovers the degree-normalized naive walk. Because each column is normalized over a different node's edges, the resulting transition probabilities remain directional ($p_{ji} \neq p_{ij}$) despite the underlying log-weights being symmetric. We enforce this symmetry because it serves as a structural regularizer that empirically yields better reproducibility than independently parameterized directed weights.

\medskip
\textbf{Link phases.} Each edge carries a phase
$\varphi_{ij} = \varphi_{ji} \in [-\pi, \pi]$ entering the walk states $|\psi_i\rangle$, with $\varphi \equiv 0$ recovering the standard walk states.

\medskip
\textbf{Node phases.} Each node carries a phase rotation $\theta_i = \pi + \delta_i$, and we train the offsets $\delta_i \in [-\pi, \pi]$. At $\delta \equiv 0$ the reflection reduces to the standard reflection of the walk.

\medskip
$U(\Theta)$ is unitary for \emph{every} value of every parameter in $\Theta = (\omega, \varphi, \delta)$: training moves through the space of legitimate quantum walks and never leaves it. The whole construction runs in $O(m)$ per step in an edge-list representation \citep{ortega2026sqwlib} --- a per-step cost that is optimal, since $m$ numbers are the
minimal description of the graph itself.

\subsection{Coherent-resolvent readout}
\label{sec:readout}

Starting from the seed's walk state $|\psi_s\rangle$ in (\ref{eq:walkop}), we accumulate the walk's evolution under a geometric discount and read out node scores by measuring the overlap with the position states $|\psi_v\rangle, v \in V$:
\begin{equation}
|\chi_s\rangle \;=\; \sum_{k=0}^{K} \gamma^k\, U^k |\psi_s\rangle ,
\qquad
x_v \;=\; \big| \langle \psi_v |\chi_s\rangle \big|^2 ,
\label{eq:readout}
\end{equation}
where $K$ is fixed by $\gamma$ so that the discarded tail is
numerically negligible, $\gamma^K \le 10^{-5}$, giving $K = 40,\,109$ for the two discounts we study, $\gamma \in \{0.7,\,
0.9\}$. We use $\gamma = 0.9$ throughout the paper, and analog results for $\gamma = 0.7$ can be found in Appendix \ref{sec:gamma}.

\medskip
\paragraph{Coherent.} Amplitudes over all walk lengths are summed \emph{first}, and the squared-norm is applied \emph{once}, at the end. The alternative order, $x_v = \sum_k \gamma^k |a_k(v)|^2$ with $a_k(v) = \langle \psi_v | U^k \psi_s\rangle$, depends on the moduli alone. Summing first also keeps the cross-length products $a_j(v)\overline{a_k(v)}$, so walks of different lengths arriving at $v$ interfere and the learned phases enter the score.

\medskip
\paragraph{Resolvent.} The geometric discount replaces a hard walk-length window, which weights all steps equally up to an abrupt cutoff, with a weighting that decay smoothly with walk length; empirically, this yields a much smoother training objective. It sets the effective horizon $1/(1-\gamma)$, the amplitude analog of a restart probability. We choose $\gamma$ so that this horizon stays \emph{below} the graph's mixing time so the walk never equilibrates within the readout and $x_v$ retains the seed's identity rather than relaxing to a degree-dominated pattern.

\medskip
\paragraph{Disclosure.} Accumulating $|\chi_s\rangle$ across walk lengths is not norm-preserving, and the readout on the states $|\psi_v\rangle$ is not a physical measurement, so the scores are not probabilities---they are real, non-negative numbers that order the nodes, which is all a ranking requires. VQIL is a \textbf{quantum-inspired classical algorithm}: unitary quantum-walk propagation at every step, with a readout designed for classical execution, and no quantum implementation claimed or required.

\subsection{VQIL training}
\label{sec:training}

\textbf{Variants.} A VQIL configuration is a choice of which parameter sets are trainable. We study six variants. Both link phases and log-weights are always parameterized symmetrically across the two directions of an undirected edge.

\noindent
\begin{minipage}[t]{0.39\linewidth}
\vspace{0pt}
\begin{enumerate}[leftmargin=*,itemsep=4pt,topsep=0pt]
\item \textbf{Naive-VQIL.} Nothing trained. The walk runs on the unweighted, phase-free graph; this is the untrained floor.
\item \textbf{Standard-VQIL.} Edge weights $\omega$ only.
\item \textbf{Link-phased-VQIL.} Edge weights $\omega$ and link phases $\varphi$.
\item \textbf{Vertex-phased-VQIL.} Edge weights $\omega$ and node phases $\delta$.
\item \textbf{Graph-phased-VQIL.} Edge weights $\omega$, link phases $\varphi$ and node phases $\delta$ together.
\item \textbf{Phased-only-VQIL.} Phases $\varphi$ and $\delta$ only, with weights $\omega$ frozen at the unweighted graph.

\end{enumerate}
\end{minipage}\hfill
\begin{minipage}[t]{0.585\linewidth}
\vspace{0pt}
\begin{algorithm}[H]
\caption{VQIL training step (graph-phased)}
\label{alg:vqil}
\renewcommand{\baselinestretch}{1.35}\normalsize
\begin{algorithmic}[1]
\STATE \textbf{Input:} edge list ($m$ directed edges); query batch
$\{(s^b, y_s^b)\}_{b=1}^{B}$; $\Theta = (\omega, \varphi, \delta)$
\STATE $p \leftarrow \mathrm{softmax}(\omega)$ over incident edges \hfill $\triangleright$~(\ref{eq:weights})
\STATE $\psi_e \leftarrow e^{i\varphi_e}\sqrt{p_e}$\ \ for every directed edge $e$ \hfill $\triangleright$~(\ref{eq:walkop})
\STATE $|\chi^b\rangle \leftarrow |\xi^b\rangle \leftarrow |\psi_{s}^b\rangle$ \hfill $\triangleright$ $|\xi^b\rangle$: running walk state
\STATE \textbf{for} $k = 1 \dots K$ \textbf{do}
\STATE \quad $|\xi^b\rangle \leftarrow S\,R(\Theta)\,|\xi^b\rangle$ \hfill $\triangleright$~(\ref{eq:walkop}), $O(m)$
\STATE \quad $|\chi^b\rangle \leftarrow |\chi^b\rangle + \gamma^k |\xi^b\rangle$ \hfill $\triangleright$ coherent sum
\STATE $x_v^b \leftarrow |\langle \psi_v|\chi^b\rangle|^2$\ \ for all $v$ \hfill $\triangleright$~(\ref{eq:readout})
\STATE $\mathcal{L} \leftarrow \frac{1}{B}\sum_b \big(1 - \rho_\tau(\tilde x_s^b, y_s^b)\big)$ \hfill $\triangleright$~(\ref{eq:loss})
\STATE $\Theta \leftarrow \mathrm{AdamW}(\Theta, \nabla_{\Theta} \mathcal{L})$;\ \ clamp
$\omega \in [-2,2]$; $\varphi, \delta \in [-\pi, \pi]$
\end{algorithmic}
\renewcommand{\baselinestretch}{1.0}\normalsize
\end{algorithm}
\end{minipage}

\vspace{6pt}
\medskip
\textbf{Loss.} For a query $(s, y_s)$, let $\tilde x_s$ denote the vector of log-transformed,
standardized scores over the $|V| - 1$ non-seed nodes. We minimize a differentiable rank
correlation, the soft-Spearman objective
\begin{equation}
\mathcal{L}(s) \;=\; 1 - \rho_\tau\!\big(\tilde x_s,\; y_s\big),
\label{eq:loss}
\end{equation}
where $\rho_\tau$ is the Spearman correlation computed with soft ranks at temperature $\tau = 1$
\citep{blondel2020}, so training never thresholds the graded labels.

\medskip
\textbf{Backpropagation.} Gradients flow exactly, by automatic differentiation, through the entire construction: the softmax normalization (\ref{eq:weights}), the complex phases in (\ref{eq:walkop}), the $K$ applications of $U$, and the squared-norm readout (\ref{eq:readout}). Algorithm~\ref{alg:vqil} summarizes one training step.

\medskip
\textbf{Protocol.} AdamW (learning rate $0.05$), up to 300 epochs with early stopping on validation Spearman correlation (patience 10), model selection at the best validation checkpoint, float64/complex128 throughout. Weights initialize at $0$ (the naive walk) and phases uniformly in $[-0.3, 0.3]$. Splits are seed-disjoint: a query's seed appears in exactly one of train/validation/test (Appendix Table~\ref{tab:datasets}).

\subsection{Baselines}
\label{sec:baselines}

\textbf{Classical control.} The matched classical control is \textbf{random walk with restart (RWR)} with learned edge weights: identical weight parameterization (\ref{eq:weights}), identical loss, optimizer, splits and early stopping, with only the propagation differing (probability-level restart iteration in place of the unitary walk). Its restart probability is fixed at $r = 1-\gamma$ so that both models weight walks of length $k$ by $\gamma^k$. We report $\gamma = 0.9$ ($r = 0.1$) as the primary block and $\gamma = 0.7$ ($r = 0.3$) alongside. A second arm, \textbf{RWR-naive} (uniform weights, $r = 1-\gamma$), marks the untrained propagation floor, exactly as Naive-VQIL does for the quantum walk.

\medskip
\textbf{Neural baselines.} We compare against three message-passing architectures: GCN \citep{kipf2017}, GAT \citep{velickovic2018} and $O(2)$ neural sheaf diffusion \citep{bodnar2022}. All three use width 64 (sheaf: 16 channels on 2-dimensional stalks), residual/diffusion updates, LayerNorm, LeakyReLU(0.1), and dropout 0.2. GAT uses single-head attention with self-loops, and the sheaf network predicts each edge's rotation angle with a per-layer MLP on the incident features. Our graphs lack node attributes, therefore we explicitly provide topological features capturing each node's relation to the query seed \citep{zhang2018seal,li2020distance,zhang2021labeling}: a seed indicator, normalized and log degrees, one-hot shortest-path distance to the seed (up to $\ge 13$), a 16-step random-walk structural encoding \citep{dwivedi2022}, and the first 16 non-trivial Laplacian eigenvectors. Node representations $h_v$ are scored by a two-layer MLP on $[\,h_v,\;h_s,\;h_v\odot h_s\,]$, $s$ the seed, followed by a softplus to keep the score positive for the ranking loss. Learning rate $\{0.005, 0.01, 0.05\}$ and depth $\{3, 6\}$ were tuned per architecture and domain (Appendix Table~\ref{tab:neural_config}). The remaining training protocol exactly matches VQIL.

\section{Experiments}
\label{sec:experiments}

\subsection{Set-up}
\label{sec:setup}

We evaluate seed-conditioned perturbation response ranking in two domains chosen so that the response is generated by a physical or biological mechanism withheld from the model, and to demonstrate scalability across orders of magnitude in graph size. For explicit details about datasets and trained models see Appendix \ref{sec:datasets}.

\medskip
\textbf{Domain 1: AC power-flow perturbation on transmission grids.} A transmission grid joins \emph{buses} (substations) by lines and transformers. At equilibrium each bus carries a voltage magnitude $V$ and a phase angle $\theta$. One query: solve AC power flow on the grid as shipped, inject a fixed load (+20\,MW, +6\,MVAr) at a seed bus, re-solve, and rank every other bus by its response. Each grid gives two instances with different physics: the angle response $|\Delta\theta|$ is real-power driven and path-mediated, whereas the voltage response $|\Delta V|$ is reactive-power-mediated and shaped by voltage regulation, since buses held at a set point by generators absorb the disturbance. The model sees only the unweighted bus graph and the seed index; reactances, transformer positions, generator identities, the slack bus, and all loads are withheld. We use six standard test networks with \textbf{30--1{,}354 nodes and 41--1{,}710 edges} (\citealp{christie1993pstca,alsac1974,vittal1992,josz2016,fliscounakis2013,zimmerman2011,thurner2018}), each partitioned seed-disjointly into five folds (60/20/20 train/validation/test, each query is tested exactly once).

\medskip
\textbf{Domain 2: In silico gene perturbation on protein--protein interaction (PPI) networks.} A pooled CRISPR screen with single-cell readout perturbs one gene per cell and reports the whole transcriptome. One query: knock down the seed gene and rank every other gene by its response---$|\mathrm{logFC}|$ of mean log-normalized expression against the pooled control cells, used as a graded relevance. Five published screens \citep{replogle2022,frangieh2021,tian2021,adamson2016}, obtained in processed form from scPerturBench \citep{scperturbench2025}, supply the instances and span cell type, CRISPR modalities, and scales (75--1{,}615 perturbations). Each screen has its own graph: its gene panel (the 5{,}000 most variable genes plus every perturbed gene) mapped onto STRING v12.0 \citep{szklarczyk2023} at medium confidence ($\ge 400$), largest component only (22--144 panel genes per dataset and five seeds in total fall outside it and are dropped rather than reconnected), giving \textbf{2{,}965--5{,}787 nodes and 29{,}059--218{,}844 edges}. The model sees only this unweighted graph and the seed index, seed-disjoint five-fold partitions as in Domain~1.

\medskip
\textbf{Metrics and protocol.} Primary metrics: Spearman $\rho$ between the predicted and
measured response over all candidates, and AUPRC@10\%, taking the strongest-responding 10\% of
candidates as positives. Secondary: AUROC@10\% on the same positive set, and NDCG@10\% with the
measured response as graded relevance. Every arm selects its epoch on validation $\rho$ and is
trained under the identical protocol of \S\ref{sec:training}. Values are mean $\pm$ sd over
evaluation cells (one task instance $\times$ one seed-disjoint fold) and arms are compared paired within
cells by a Wilcoxon signed-rank test.

\subsection{Results}
\label{sec:results}

\textbf{Robust generalization across graph topologies and scales.} On the six transmission grids (Table~\ref{tab:grid}), weight training alone improves Spearman $\rho$ from $0.312/0.147$ (angle/voltage) to $0.880/0.799$, with the best phase-bearing variants reaching $0.917/0.831$. On the PPI networks (Table~\ref{tab:ppi}), up to two orders of magnitude larger in scale, $\rho$ lifts from $0.171$ to $0.620$. Among the trainable axes, edge weights drive the majority of the gain. Freezing the weights and training only the phases (Phased-only-VQIL) trails weights-only training by $0.128$ (angle) and $0.211$ (voltage) on the power grids and by $0.113$ on the PPI networks. Per-node phases add a smaller but consistent gain on top of weights: $+0.031$ (angle) and $+0.032$ (voltage) in $\rho$ on the power grids (27/30 and 28/30 cells, $p\le2.4{\times}10^{-6}$; AUPRC $+0.080$ and $+0.052$) and $+0.033$ on the PPI networks. Link phases contribute no measurable benefit: Link-phased-VQIL is indistinguishable from Standard-VQIL ($|\Delta\rho| \leq 0.0004$, $p\ge0.57$), and Graph-phased-VQIL mirrors Vertex-phased-VQIL ($|\Delta\rho| \leq 0.006$). We therefore treat edge weights and node phases as the two effective trainable axes and adopt Vertex-phased-VQIL as the main variant. 

\medskip
Compared with the matched classical control, the unitary walk demonstrates a clear post-training advantage. Prior to training, the classical walk ranks better on unweighted graphs across all three blocks, indicating the unitary walk holds no intrinsic zero-shot advantage. Post-training, this ordering decisively reverses: Vertex-phased-VQIL outperforms the classical control by $0.134$ and $0.149$ in $\rho$ on the power grids, and by $0.050$ on the PPI networks. 

\medskip
Finally, we benchmark against seed-conditioned neural baselines, which benefit from additional structural node features and a pair decoder. On the power grids, Vertex-phased-VQIL matches GCN and Sheaf-$\mathcal{O}(2)$ on the angle response, falls $0.020$ below GAT, and trails all three on the voltage response. On the PPI networks the order reverses: Vertex-phased-VQIL outperforms GCN, GAT and Sheaf-$\mathcal{O}(2)$ by $0.038$, $0.091$ and $0.255$ in $\rho$ (25/25 cells, $p\le6{\times}10^{-8}$), leading GCN, the strongest neural baseline, on every dataset and on all four metrics. GCN in turn matches or exceeds the classical control. VQIL is also cheaper to train: $2.4\times$ to $4.6\times$ less total wall-clock time on the power grids and $3.8\times$ to $7.3\times$ less on each PPI network (Appendix Table~\ref{tab:time}). Its edge-indexed parameterization uses $4.9\times$ to over $400\times$ fewer parameters than the neural baselines on the grids but grows with graph size, reaching $0.8\times$ to $15\times$ their size on the PPI networks (Appendix Table~\ref{tab:params}).

\begin{table}[!ht]
\centering
\caption{Power-grid perturbation-response ranking. Mean $\pm$ sd over 30 cells (six power grids $\times$ five seed-disjoint folds). VQIL variants use symmetric edge weights and link phases with discount $\gamma=0.9$; the classical control uses the matched restart $r=1-\gamma=0.1$. Results at $\gamma = 0.7$ are in Appendix Table~\ref{tab:grid_g07}. \textbf{Bold}: main VQIL variant.}
\label{tab:grid}
\vspace{1ex} %
\small
\begin{tabular*}{\linewidth}{@{\extracolsep{\fill}} l cc cc @{}}
\toprule
& \multicolumn{2}{c}{Phase-angle response $|\Delta\theta|$}
& \multicolumn{2}{c}{Voltage response $|\Delta V|$} \\
\cmidrule(lr){2-3}\cmidrule(lr){4-5}
Model & Spearman $\rho$ & AUPRC@10\% & Spearman $\rho$ & AUPRC@10\% \\
\midrule
Naive-VQIL (untrained) & 0.312 $\pm$ 0.269 & 0.541 $\pm$ 0.210 & 0.147 $\pm$ 0.154 & 0.310 $\pm$ 0.074 \\
Standard-VQIL & 0.880 $\pm$ 0.049 & 0.750 $\pm$ 0.107 & 0.799 $\pm$ 0.066 & 0.573 $\pm$ 0.132 \\
Link-phased-VQIL & 0.881 $\pm$ 0.048 & 0.749 $\pm$ 0.108 & 0.799 $\pm$ 0.066 & 0.572 $\pm$ 0.133 \\
\textbf{Vertex-phased-VQIL} & \textbf{0.911 $\pm$ 0.043} & \textbf{0.830 $\pm$ 0.062} & \textbf{0.831 $\pm$ 0.056} & \textbf{0.624 $\pm$ 0.144} \\
Graph-phased-VQIL & 0.917 $\pm$ 0.042 & 0.835 $\pm$ 0.065 & 0.829 $\pm$ 0.056 & 0.623 $\pm$ 0.144 \\
Phased-only-VQIL & 0.752 $\pm$ 0.144 & 0.757 $\pm$ 0.101 & 0.588 $\pm$ 0.096 & 0.545 $\pm$ 0.156 \\
\midrule
Naive-RWR (untrained) & 0.411 $\pm$ 0.170 & 0.662 $\pm$ 0.109 & 0.228 $\pm$ 0.110 & 0.407 $\pm$ 0.114 \\
Trained-RWR & 0.777 $\pm$ 0.091 & 0.744 $\pm$ 0.083 & 0.681 $\pm$ 0.090 & 0.544 $\pm$ 0.140 \\
\midrule
GCN & 0.915 $\pm$ 0.071 & 0.813 $\pm$ 0.079 & 0.920 $\pm$ 0.041 & 0.789 $\pm$ 0.102 \\
GAT & 0.932 $\pm$ 0.058 & 0.830 $\pm$ 0.076 & 0.920 $\pm$ 0.041 & 0.795 $\pm$ 0.107 \\
Sheaf-$\mathcal{O}(2)$ & 0.913 $\pm$ 0.081 & 0.796 $\pm$ 0.095 & 0.908 $\pm$ 0.039 & 0.735 $\pm$ 0.116 \\
\bottomrule
\end{tabular*}
\end{table}

\begin{table}[!ht]
\centering
\caption{Gene-perturbation response ranking on protein--protein interaction graphs. Mean $\pm$ sd over 25 cells (five Perturb-seq datasets $\times$ five seed-disjoint folds). Model configurations match Table~\ref{tab:grid} ($\gamma=0.9$). Results at $\gamma = 0.7$ are in Appendix Table~\ref{tab:ppi_g07}. \textbf{Bold}: main VQIL variant.}
\label{tab:ppi}
\vspace{1ex} %
\small
\resizebox{0.90\linewidth}{!}{%
\begin{tabular}{l cccc}
\toprule
& \multicolumn{4}{c}{Gene expression response $|\mathrm{logFC}|$} \\
\cmidrule(lr){2-5}
Model & Spearman $\rho$ & AUPRC@10\% & AUROC@10\% & NDCG@10\% \\
\midrule
Naive-VQIL (untrained) & 0.171 $\pm$ 0.050 & 0.155 $\pm$ 0.026 & 0.616 $\pm$ 0.040 & 0.330 $\pm$ 0.053 \\
Standard-VQIL & 0.587 $\pm$ 0.157 & 0.336 $\pm$ 0.080 & 0.824 $\pm$ 0.039 & 0.532 $\pm$ 0.045 \\
Link-phased-VQIL & 0.587 $\pm$ 0.157 & 0.336 $\pm$ 0.080 & 0.824 $\pm$ 0.039 & 0.532 $\pm$ 0.044 \\
\textbf{Vertex-phased-VQIL} & \textbf{0.620 $\pm$ 0.167} & \textbf{0.361 $\pm$ 0.086} & \textbf{0.849 $\pm$ 0.042} & \textbf{0.552 $\pm$ 0.045} \\
Graph-phased-VQIL & 0.620 $\pm$ 0.167 & 0.361 $\pm$ 0.085 & 0.849 $\pm$ 0.042 & 0.554 $\pm$ 0.047 \\
Phased-only-VQIL & 0.474 $\pm$ 0.108 & 0.270 $\pm$ 0.054 & 0.784 $\pm$ 0.037 & 0.468 $\pm$ 0.054 \\
\midrule
\makecell[l]{Naive-RWR (untrained)} & 0.174 $\pm$ 0.055 & 0.157 $\pm$ 0.029 & 0.617 $\pm$ 0.042 & 0.335 $\pm$ 0.042 \\
\makecell[l]{Trained-RWR} & 0.570 $\pm$ 0.150 & 0.296 $\pm$ 0.073 & 0.808 $\pm$ 0.041 & 0.491 $\pm$ 0.047 \\
\midrule
GCN & 0.582 $\pm$ 0.166 & 0.314 $\pm$ 0.077 & 0.815 $\pm$ 0.047 & 0.518 $\pm$ 0.056 \\
GAT & 0.528 $\pm$ 0.183 & 0.285 $\pm$ 0.088 & 0.782 $\pm$ 0.063 & 0.487 $\pm$ 0.070 \\
Sheaf-$\mathcal{O}(2)$ & 0.365 $\pm$ 0.141 & 0.208 $\pm$ 0.039 & 0.696 $\pm$ 0.043 & 0.411 $\pm$ 0.032 \\
\bottomrule
\end{tabular}}
\end{table}

\medskip
\textbf{The coherent resolvent readout enables smooth training.} To isolate what the resolvent contributes, we factor the readout into two independent choices: how walk lengths are weighted (geometric $\gamma^k$ versus a uniform window) and when the squared-norm is applied (amplitudes summed then squared, versus each step squared then summed). We train all four combinations at two horizons ($\gamma=0.7, 0.9$) from identical per-step amplitudes (Table~\ref{tab:readout};Appendix Table~\ref{tab:readout07}). At $\gamma=0.9$ the coherent geometric readout used by VQIL is the best of the four in every task: it leads its square-then-add counterpart (order) by $+0.103$ in $\rho$ and the uniform window (weighting) by $+0.105$ on the PPI networks, and by $+0.072$/$+0.098$ (order) and $+0.208$/$+0.057$ (weighting) on the power grids (angle/voltage). Of the two geometric readouts, it is also the only readout that benefits from the longer horizon: moving from $\gamma=0.7$ to $0.9$ lifts it by $+0.063$ and $+0.127\,\rho$ on the grids and $+0.033\,\rho$ on the PPI networks, whereas square-then-add changes by $-0.021$ and $+0.018$ on the grids and $-0.019$ on the PPI networks. The benefit of coherent summation grows from $-0.012$ to $+0.072$ in $\rho$ on the grid angle response and from $+0.051$ to $+0.103$ on the PPI networks. 

\medskip
The two components of the readout contribute to trainability in distinct ways, which we measure directly on the loss surface. Evaluating all four readouts at common parameter points (the untrained walk and the trained Vertex-phased-VQIL solution) along shared random directions, the geometric discount is what makes the objective smooth: a uniform window of the same horizon raises the curvature of the loss (mean absolute second difference relative to its range) by $22\times$ on the grids and $18\times$ on the PPI networks under coherent summation, and by $2.4\times$ to $6.4\times$ under square-then-add. Coherent summation then smooths the surface further, around trained solutions: $1.3\times$ lower curvature on the grids and $1.6\times$ on the PPI networks, and $1.7\times$ and $2.4\times$ lower detrended total variation. On the grids it also makes the gradient consistent from the outset: gradients computed on disjoint minibatches at the untrained walk agree with median cosine $0.94$ versus $0.54$ for square-then-add. The coherent readout is therefore not an accuracy trade-off against trainability: its geometric weighting yields a smooth objective, and its coherence makes the longer horizon informative while producing a better-conditioned surface around the solutions it finds.

\begin{table}[!ht]
\centering
\caption{Readout ablation at the $\gamma=0.9$ horizon: the $2\times2$ factorial over step weighting ($K = 109$) and the order in which the squared norm is taken. All arms are Vertex-phased-VQIL trained under the protocol of Tables~\ref{tab:grid} and \ref{tab:ppi}. Spearman $\rho$, mean $\pm$ sd over five seed-disjoint folds of six grids (30 cells) and five Perturb-seq datasets (25 cells). \textbf{Bold}: coherent resolvent readout of VQIL.}
\label{tab:readout}
\vspace{1ex}
\small
\begin{tabular*}{\linewidth}{@{\extracolsep{\fill}} l cc c @{}}
\toprule
& \multicolumn{2}{c}{Power grids} & PPI networks \\
\cmidrule(lr){2-3}\cmidrule(lr){4-4}
Readout & Phase-angle $|\Delta\theta|$ & Voltage $|\Delta V|$ & Gene expression \\
\midrule
Uniform, add-then-square & 0.703 $\pm$ 0.165 & 0.773 $\pm$ 0.099 & 0.515 $\pm$ 0.136 \\
Uniform, square-then-add & 0.782 $\pm$ 0.191 & 0.692 $\pm$ 0.149 & 0.453 $\pm$ 0.155 \\
\textbf{Geometric, add-then-square} & \textbf{0.911 $\pm$ 0.043} & \textbf{0.831 $\pm$ 0.056} & \textbf{0.620 $\pm$ 0.167} \\
Geometric, square-then-add & 0.839 $\pm$ 0.126 & 0.733 $\pm$ 0.093 & 0.517 $\pm$ 0.156 \\
\bottomrule
\end{tabular*}
\end{table}

\begin{figure}[!ht]
\centering
\includegraphics[width=0.95\linewidth]{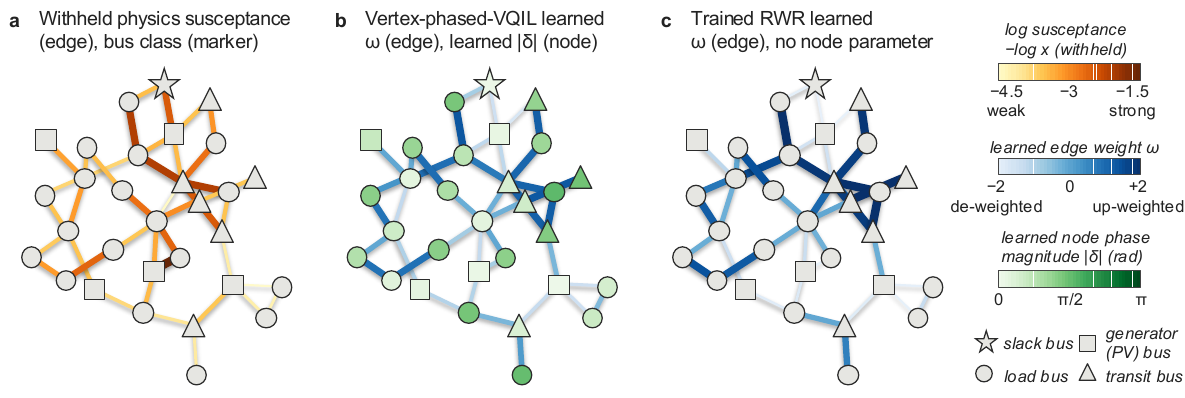}
\caption{Trained fields on the 30-bus grid (case30; voltage response, $\gamma = 0.9$, five-fold means). (a) Ground-truth branch susceptance (edges) and bus class (markers). (b) Vertex-phased-VQIL: edge weights ($\omega$), which follow susceptance, and node-phase magnitudes $|\delta|$, smallest at regulated buses. (c) Trained-RWR learns only edge weights ($\omega$). Fields show 5-fold means.}
\label{fig:visibility}
\end{figure}

\medskip
\textbf{The trained fields read back as statements about the withheld mechanism.} Every parameter sits on a named edge or node (Definition~\ref{def:visibility}): the fitted Vertex-phased-VQIL is one weight $\omega \in [-2,2]$ per edge and one phase $\delta \in [-\pi, \pi]$ per node, readable without post hoc attribution. Neural baselines have no such correspondence, and trained-RWR carries weights only. We therefore compare the classical and quantum $\omega$ fields on identical edges and evaluate VQIL's unique $\delta$ field against withheld annotations: branch susceptance and bus type on the power grids, and co-expression and the response propensity on the PPI network (Figure~\ref{fig:visibility}, Appendix Figure~\ref{fig:grid-vis}, Appendix Figure~\ref{fig:ppi-vis}). Both fields are reproducible: across training folds, $\omega$ correlates at $\rho=0.68$--$0.96$ and $|\delta|$ at $\rho=0.37$--$0.83$. The phase field is broadly distributed (median $|\delta|\in[0.24,0.90]$\,rad) with no boundary saturation (Appendix Figure~\ref{fig:delta_distribution}). The sign of $\delta$, however, does not reproduce across folds (mean cross-fold $|\rho|<0.03$). A simultaneous flip of all node phases ($\delta\to-\delta$) leaves the readout exactly invariant, so the overall sign is unidentifiable. Relative signs are not constrained by this symmetry but are empirically unstable, so we interpret $|\delta|$ only.

\medskip
\textbf{Power grids.} A bus perturbation propagates in proportion to branch susceptance and is arrested at voltage-regulated nodes. The VQIL and RWR weight fields largely agree (Appendix Figure~\ref{fig:omega_correlation}), but RWR drives about twice as many weights to the $\pm2$ boundary ($46\%$ vs.\ $21\%$ of edges; Appendix Figure~\ref{fig:omega_distribution}). VQIL's weights also carry susceptance information beyond RWR's: after partialling out the RWR weights, the VQIL residual still correlates with withheld log-susceptance ($\rho = 0.06$, $p = 3\times10^{-3}$), whereas the RWR residual does not ($\rho = -0.01$, n.s.). The phase magnitudes capture voltage regulation $|\delta|$: it is lowest at voltage-regulated buses, and separates generator buses from all other buses with AUC $0.74$ (degree-matched null $0.56$).

\medskip
\textbf{PPI networks.} Perturbations propagate through physical interactions modulated by cellular co-regulation. The two weight fields again agree closely (Appendix Figure~\ref{fig:omega_correlation}), with RWR saturating more edges ($43\%$ vs.\ $25\%$), and both track control-cell co-expression to a similar, modest degree (VQIL $\rho = 0.11$--$0.28$; RWR $0.11$--$0.27$). The phase field, which has no RWR counterpart, is larger at genes that respond strongly across many perturbations, capturing stress-responsive genes (AUC $0.71$--$0.87$), independently of their baseline expression level. Across both domains, the learned weights encode transmission strength and the phase magnitudes track local response propensity: low at regulated buses, high at strongly responsive genes.

\section{Discussion and Conclusion}
\label{sec:discussion}

\medskip
In this work, we introduced VQIL, a scalable framework taking advantage of quantum interference as an alternative to conventional graph representation learning. Standard random walks and graph neural networks frequently struggle with oversmoothing, where information homogenizes around high-degree topological hubs. By introducing learned local node phases ($\delta$), VQIL leverages interference to dynamically bypass these bottlenecks, preserving critical local signals such as responsive gene modules or regulated electrical buses. A coherent resolvent readout highlights that amplitude superposition is not merely an incidental feature, but the foundational mechanism enabling targeted signal routing.

\medskip
Beyond competitive empirical performance, VQIL's primary contribution lies in mechanistic visibility. Deep neural baselines obscure predictive logic within dense, distributed latent spaces, whereas classical walks restrict interpretability exclusively to edge transitions. In contrast, VQIL maps every scalar parameter directly to an identifiable physical component. Edge weights ($\omega$) denote primary transmission corridors, while node-phase magnitudes ($\vert{}\delta\vert{}$) mark explicit sinks where perturbations are halted. This turns the trained model into an interpretable hypothesis, which is vital for discovery-driven fields like computational biology.

\medskip
Despite these strengths, several architectural limitations define clear future directions. Currently, VQIL remains a transductive ranking tool bound to fixed topologies, incapable of predicting signed, continuous expression values or generalizing to unseen graphs. Future extensions could introduce lightweight encoders to parameterize $\omega$ and $\delta$ inductively from features. Furthermore, moving beyond symmetric constraints to incorporate anti-symmetric link phases in directed networks will unlock causal modeling where detailed balance is fundamentally broken.

\section*{Funding}
This work was supported by the Korean ARPA-H Project through the Korea Health Industry Development Institute (KHIDI), funded by the Ministry of Health \& Welfare, Korea (RS-2025-25456722); the Brain Pool Plus Fellowship Program, funded by the Ministry of Science and ICT (RS-2025-25427881); the Institute of Information \& Communications Technology Planning \& Evaluation (IITP)-ITRC (Information Technology Research Center) grant funded by the Korea government (IITP-2026-RS-2026-25519864); the National Research Foundation of Korea (RS-2025-02309510); the Ministry of Trade, Industry, and Energy (MOTIE), Korea, under the Industrial Innovation Infrastructure Development Project (RS-2024-00466693); and the InnoCORE program of the Ministry of Science and ICT (PICORE Center, N10260101). S.L.'s PhD studentship is jointly funded by the Cambridge Stem Cell Institute and the Milner Therapeutics Institute.

\bibliography{vqil}

@inproceedings{szegedy2004,
  author    = {Szegedy, Mario},
  title     = {Quantum Speed-Up of {M}arkov Chain Based Algorithms},
  booktitle = {Proceedings of the 45th Annual IEEE Symposium on Foundations of Computer Science (FOCS)},
  pages     = {32--41},
  year      = {2004},
  doi       = {10.1109/FOCS.2004.53}
}

@inproceedings{aharonov2001,
  author    = {Aharonov, Dorit and Ambainis, Andris and Kempe, Julia and Vazirani, Umesh},
  title     = {Quantum Walks on Graphs},
  booktitle = {Proceedings of the 33rd Annual ACM Symposium on Theory of Computing (STOC)},
  pages     = {50--59},
  year      = {2001},
  doi       = {10.1145/380752.380758}
}

@article{paparo2012,
  author  = {Paparo, Giuseppe Davide and Martin-Delgado, Miguel Angel},
  title   = {{G}oogle in a Quantum Network},
  journal = {Scientific Reports},
  volume  = {2},
  pages   = {444},
  year    = {2012},
  doi     = {10.1038/srep00444}
}

@article{paparo2013,
  author  = {Paparo, Giuseppe Davide and M{\"u}ller, Markus and Comellas, Francesc and Martin-Delgado, Miguel Angel},
  title   = {Quantum {G}oogle in a Complex Network},
  journal = {Scientific Reports},
  volume  = {3},
  pages   = {2773},
  year    = {2013},
  doi     = {10.1038/srep02773}
}

@article{ortega2023,
  author  = {Ortega, Sergio A. and Martin-Delgado, Miguel A.},
  title   = {Generalized Quantum {P}age{R}ank Algorithm with Arbitrary Phase Rotations},
  journal = {Physical Review Research},
  volume  = {5},
  number  = {1},
  pages   = {013061},
  year    = {2023},
  doi     = {10.1103/PhysRevResearch.5.013061}
}

@article{ortega2024squwals,
  author  = {Ortega, Sergio A. and Martin-Delgado, Miguel A.},
  title   = {{SQUWALS}: A {S}zegedy {QU}antum {WAL}ks Simulator},
  journal = {Advanced Quantum Technologies},
  volume  = {7},
  number  = {7},
  pages   = {2400022},
  year    = {2024},
  doi     = {10.1002/qute.202400022}
}

@article{ortega2025,
  author  = {Ortega, Sergio A. and Martin-Delgado, Miguel A.},
  title   = {Complex-Phase Extensions of the {S}zegedy Quantum Walk on Graphs},
  journal = {Physical Review A},
  volume  = {111},
  number  = {3},
  pages   = {032216},
  year    = {2025},
  doi     = {10.1103/PhysRevA.111.032216}
}

@misc{ortega2026sqwlib,
  author        = {Ortega, Sergio A. and Park, Daniel K.},
  title         = {Efficient Simulation of {S}zegedy Quantum Walk Formulations and Algorithms},
  year          = {2026},
  eprint        = {2606.14226},
  archivePrefix = {arXiv},
  howpublished  = {arXiv:2606.14226}
}

@inproceedings{dernbach2018,
  author    = {Dernbach, Stefan and Mohseni-Kabir, Arman and Pal, Siddharth and Towsley, Don},
  title     = {Quantum Walk Neural Networks for Graph-Structured Data},
  booktitle = {Complex Networks and Their Applications VII},
  series    = {Studies in Computational Intelligence},
  publisher = {Springer},
  pages     = {182--193},
  year      = {2018},
  doi       = {10.1007/978-3-030-05414-4_15}
}

@article{dernbach2019,
  author  = {Dernbach, Stefan and Mohseni-Kabir, Arman and Pal, Siddharth and Gepner, Miles and Towsley, Don},
  title   = {Quantum Walk Neural Networks with Feature Dependent Coins},
  journal = {Applied Network Science},
  volume  = {4},
  number  = {1},
  pages   = {76},
  year    = {2019},
  doi     = {10.1007/s41109-019-0188-2}
}

@article{saarinen2024,
  author  = {Saarinen, Harto and Goldsmith, Mark and Wang, Rui-Sheng and Loscalzo, Joseph and Maniscalco, Sabrina},
  title   = {Disease Gene Prioritization with Quantum Walks},
  journal = {Bioinformatics},
  volume  = {40},
  number  = {8},
  pages   = {btae513},
  year    = {2024},
  doi     = {10.1093/bioinformatics/btae513}
}

@article{park2026longcovid,
  author  = {Park, Jaesub and Hwang, Woochang and Lee, Seokjun and Lee, Hyun Chang and MacMahon, M{\'e}abh and Zilbauer, Matthias and Han, Namshik},
  title   = {Advancing Understanding of Long {COVID} Pathophysiology through Quantum Walk-Based Network Analysis},
  journal = {Bioinformatics Advances},
  volume  = {6},
  number  = {1},
  pages   = {vbag050},
  year    = {2026},
  doi     = {10.1093/bioadv/vbag050}
}

@techreport{page1999,
  author      = {Page, Lawrence and Brin, Sergey and Motwani, Rajeev and Winograd, Terry},
  title       = {The {P}age{R}ank Citation Ranking: Bringing Order to the Web},
  institution = {Stanford InfoLab},
  number      = {1999-66},
  year        = {1999}
}

@inproceedings{tong2006,
  author    = {Tong, Hanghang and Faloutsos, Christos and Pan, Jia-Yu},
  title     = {Fast Random Walk with Restart and Its Applications},
  booktitle = {Proceedings of the 6th IEEE International Conference on Data Mining (ICDM)},
  pages     = {613--622},
  year      = {2006},
  doi       = {10.1109/ICDM.2006.70}
}

@article{cowen2017,
  author  = {Cowen, Lenore and Ideker, Trey and Raphael, Benjamin J. and Sharan, Roded},
  title   = {Network Propagation: A Universal Amplifier of Genetic Associations},
  journal = {Nature Reviews Genetics},
  volume  = {18},
  number  = {9},
  pages   = {551--562},
  year    = {2017},
  doi     = {10.1038/nrg.2017.38}
}

@article{han2021,
  author  = {Han, Namshik and Hwang, Woochang and Tzelepis, Konstantinos and Schmerer, Patrick and Yankova, Eliza and MacMahon, M{\'e}abh and Lei, Winnie and Katritsis, Nicholas M. and Liu, Anika and Felgenhauer, Ulrike and Schuldt, Alison and Harris, Rebecca and Chapman, Kathryn and McCaughan, Frank and Weber, Friedemann and Kouzarides, Tony},
  title   = {Identification of {SARS-CoV-2}--Induced Pathways Reveals Drug Repurposing Strategies},
  journal = {Science Advances},
  volume  = {7},
  number  = {27},
  pages   = {eabh3032},
  year    = {2021},
  doi     = {10.1126/sciadv.abh3032}
}

@inproceedings{agarwal2006,
  author    = {Agarwal, Alekh and Chakrabarti, Soumen and Aggarwal, Sunny},
  title     = {Learning to Rank Networked Entities},
  booktitle = {Proceedings of the 12th ACM SIGKDD International Conference on Knowledge Discovery and Data Mining (KDD)},
  pages     = {14--23},
  year      = {2006},
  doi       = {10.1145/1150402.1150409}
}

@inproceedings{backstrom2011,
  author    = {Backstrom, Lars and Leskovec, Jure},
  title     = {Supervised Random Walks: Predicting and Recommending Links in Social Networks},
  booktitle = {Proceedings of the 4th ACM International Conference on Web Search and Data Mining (WSDM)},
  pages     = {635--644},
  year      = {2011},
  doi       = {10.1145/1935826.1935914}
}

@article{jin2019,
  author  = {Jin, Woojeong and Jung, Jinhong and Kang, U},
  title   = {Supervised and Extended Restart in Random Walks for Ranking and Link Prediction in Networks},
  journal = {PLOS ONE},
  volume  = {14},
  number  = {3},
  pages   = {e0213857},
  year    = {2019},
  doi     = {10.1371/journal.pone.0213857}
}

@inproceedings{kipf2017,
  author    = {Kipf, Thomas N. and Welling, Max},
  title     = {Semi-Supervised Classification with Graph Convolutional Networks},
  booktitle = {International Conference on Learning Representations (ICLR)},
  year      = {2017}
}

@inproceedings{velickovic2018,
  author    = {Veli{\v{c}}kovi{\'c}, Petar and Cucurull, Guillem and Casanova, Arantxa and Romero, Adriana and Li{\`o}, Pietro and Bengio, Yoshua},
  title     = {Graph Attention Networks},
  booktitle = {International Conference on Learning Representations (ICLR)},
  year      = {2018}
}

@inproceedings{bodnar2022,
  author    = {Bodnar, Cristian and Di Giovanni, Francesco and Chamberlain, Benjamin Paul and Li{\`o}, Pietro and Bronstein, Michael M.},
  title     = {Neural Sheaf Diffusion: A Topological Perspective on Heterophily and Oversmoothing in {GNN}s},
  booktitle = {Advances in Neural Information Processing Systems (NeurIPS)},
  volume    = {35},
  pages     = {18527--18541},
  year      = {2022}
}

@inproceedings{dwivedi2022,
  author    = {Dwivedi, Vijay Prakash and Luu, Anh Tuan and Laurent, Thomas and Bengio, Yoshua and Bresson, Xavier},
  title     = {Graph Neural Networks with Learnable Structural and Positional Representations},
  booktitle = {International Conference on Learning Representations (ICLR)},
  year      = {2022}
}

@inproceedings{li2020distance,
  author    = {Li, Pan and Wang, Yanbang and Wang, Hongwei and Leskovec, Jure},
  title     = {Distance Encoding: Design Provably More Powerful Neural Networks for Graph Representation Learning},
  booktitle = {Advances in Neural Information Processing Systems (NeurIPS)},
  volume    = {33},
  pages     = {4465--4478},
  year      = {2020}
}

@inproceedings{zhang2018seal,
  author    = {Zhang, Muhan and Chen, Yixin},
  title     = {Link Prediction Based on Graph Neural Networks},
  booktitle = {Advances in Neural Information Processing Systems (NeurIPS)},
  volume    = {31},
  year      = {2018}
}

@inproceedings{zhang2021labeling,
  author    = {Zhang, Muhan and Li, Pan and Xia, Yinglong and Wang, Kai and Jin, Long},
  title     = {Labeling Trick: A Theory of Using Graph Neural Networks for Multi-Node Representation Learning},
  booktitle = {Advances in Neural Information Processing Systems (NeurIPS)},
  volume    = {34},
  pages     = {9061--9073},
  year      = {2021}
}

@inproceedings{blondel2020,
  author    = {Blondel, Mathieu and Teboul, Olivier and Berthet, Quentin and Djolonga, Josip},
  title     = {Fast Differentiable Sorting and Ranking},
  booktitle = {Proceedings of the 37th International Conference on Machine Learning (ICML)},
  series    = {Proceedings of Machine Learning Research},
  volume    = {119},
  pages     = {950--959},
  year      = {2020}
}

@article{ma2018,
  author  = {Ma, Jianzhu and Yu, Michael Ku and Fong, Samson and Ono, Keiichiro and Sage, Eric and Demchak, Barry and Sharan, Roded and Ideker, Trey},
  title   = {Using Deep Learning to Model the Hierarchical Structure and Function of a Cell},
  journal = {Nature Methods},
  volume  = {15},
  number  = {4},
  pages   = {290--298},
  year    = {2018},
  doi     = {10.1038/nmeth.4627}
}

@article{yu2018,
  author  = {Yu, Michael K. and Ma, Jianzhu and Fisher, Jasmin and Kreisberg, Jason F. and Raphael, Benjamin J. and Ideker, Trey},
  title   = {Visible Machine Learning for Biomedicine},
  journal = {Cell},
  volume  = {173},
  number  = {7},
  pages   = {1562--1565},
  year    = {2018},
  doi     = {10.1016/j.cell.2018.05.056}
}

@article{rudin2019,
  author  = {Rudin, Cynthia},
  title   = {Stop Explaining Black Box Machine Learning Models for High Stakes Decisions and Use Interpretable Models Instead},
  journal = {Nature Machine Intelligence},
  volume  = {1},
  number  = {5},
  pages   = {206--215},
  year    = {2019},
  doi     = {10.1038/s42256-019-0048-x}
}

@article{zimmerman2011,
  author  = {Zimmerman, Ray D. and Murillo-S{\'a}nchez, Carlos E. and Thomas, Robert J.},
  title   = {{MATPOWER}: Steady-State Operations, Planning, and Analysis Tools for Power Systems Research and Education},
  journal = {IEEE Transactions on Power Systems},
  volume  = {26},
  number  = {1},
  pages   = {12--19},
  year    = {2011},
  doi     = {10.1109/TPWRS.2010.2051168}
}

@article{thurner2018,
  author  = {Thurner, Leon and Scheidler, Alexander and Sch{\"a}fer, Florian and Menke, Jan-Hendrik and Dollichon, Julian and Meier, Friederike and Meinecke, Steffen and Braun, Martin},
  title   = {pandapower---An Open-Source {P}ython Tool for Convenient Modeling, Analysis, and Optimization of Electric Power Systems},
  journal = {IEEE Transactions on Power Systems},
  volume  = {33},
  number  = {6},
  pages   = {6510--6521},
  year    = {2018},
  doi     = {10.1109/TPWRS.2018.2829021}
}

@misc{christie1993pstca,
  author       = {Christie, Richard D.},
  title        = {Power Systems Test Case Archive},
  howpublished = {University of Washington, \url{https://labs.ece.uw.edu/pstca/}},
  year         = {1993}
}

@article{alsac1974,
  author  = {Alsac, O. and Stott, B.},
  title   = {Optimal Load Flow with Steady-State Security},
  journal = {IEEE Transactions on Power Apparatus and Systems},
  volume  = {PAS-93},
  number  = {3},
  pages   = {745--751},
  year    = {1974},
  doi     = {10.1109/TPAS.1974.293972}
}

@article{vittal1992,
  author  = {Vittal, Vijay and Martin, D. and Chu, R. and Fish, J. and Giri, J. C. and Tang, C. K. and Villaseca, F. Eugenio and Farmer, R. G.},
  title   = {Transient Stability Test Systems for Direct Stability Methods},
  journal = {IEEE Transactions on Power Systems},
  volume  = {7},
  number  = {1},
  pages   = {37--43},
  year    = {1992},
  doi     = {10.1109/59.141684}
}

@misc{josz2016,
  author        = {Josz, C{\'e}dric and Fliscounakis, St{\'e}phane and Maeght, Jean and Panciatici, Patrick},
  title         = {{AC} Power Flow Data in {MATPOWER} and {QCQP} Format: {iTesla}, {RTE} Snapshots, and {PEGASE}},
  year          = {2016},
  eprint        = {1603.01533},
  archivePrefix = {arXiv},
  howpublished  = {arXiv:1603.01533}
}

@article{fliscounakis2013,
  author  = {Fliscounakis, St{\'e}phane and Panciatici, Patrick and Capitanescu, Florin and Wehenkel, Louis},
  title   = {Contingency Ranking with Respect to Overloads in Very Large Power Systems Taking into Account Uncertainty, Preventive, and Corrective Actions},
  journal = {IEEE Transactions on Power Systems},
  volume  = {28},
  number  = {4},
  pages   = {4909--4917},
  year    = {2013},
  doi     = {10.1109/TPWRS.2013.2251015}
}

@article{replogle2022,
  author  = {Replogle, Joseph M. and Saunders, Reuben A. and Pogson, Angela N. and Hussmann, Jeffrey A. and Lenail, Alexander and Guna, Alina and Mascibroda, Lauren and Wagner, Eric J. and Adelman, Karen and Lithwick-Yanai, Gila and Iremadze, Nika and Oberstrass, Florian and Lipson, Doron and Bonnar, Jessica L. and Jost, Marco and Norman, Thomas M. and Weissman, Jonathan S.},
  title   = {Mapping Information-Rich Genotype--Phenotype Landscapes with Genome-Scale {P}erturb-seq},
  journal = {Cell},
  volume  = {185},
  number  = {14},
  pages   = {2559--2575.e28},
  year    = {2022},
  doi     = {10.1016/j.cell.2022.05.013}
}

@article{frangieh2021,
  author  = {Frangieh, Chris J. and Melms, Johannes C. and Thakore, Pratiksha I. and Geiger-Schuller, Kathryn R. and Ho, Patricia and Luoma, Adrienne M. and Cleary, Brian and Jerby-Arnon, Livnat and Malu, Shruti and Cuoco, Michael S. and Zhao, Maryann and Ager, Casey R. and Rogava, Meri and Hovey, Lila and Rotem, Asaf and Bernatchez, Chantale and Wucherpfennig, Kai W. and Johnson, Bruce E. and Rozenblatt-Rosen, Orit and Schadendorf, Dirk and Regev, Aviv and Izar, Benjamin},
  title   = {Multimodal Pooled {P}erturb-{CITE}-seq Screens in Patient Models Define Mechanisms of Cancer Immune Evasion},
  journal = {Nature Genetics},
  volume  = {53},
  number  = {3},
  pages   = {332--341},
  year    = {2021},
  doi     = {10.1038/s41588-021-00779-1}
}

@article{tian2021,
  author  = {Tian, Ruilin and Abarientos, Anthony and Hong, Jason and Hashemi, Sayed Hadi and Yan, Rui and Dr{\"a}ger, Nina and Leng, Kun and Nalls, Mike A. and Singleton, Andrew B. and Xu, Ke and Faghri, Faraz and Kampmann, Martin},
  title   = {Genome-Wide {CRISPRi/a} Screens in Human Neurons Link Lysosomal Failure to Ferroptosis},
  journal = {Nature Neuroscience},
  volume  = {24},
  number  = {7},
  pages   = {1020--1034},
  year    = {2021},
  doi     = {10.1038/s41593-021-00862-0}
}

@article{adamson2016,
  author  = {Adamson, Britt and Norman, Thomas M. and Jost, Marco and Cho, Min Y. and Nu{\~n}ez, James K. and Chen, Yuwen and Villalta, Jacqueline E. and Gilbert, Luke A. and Horlbeck, Max A. and Hein, Marco Y. and Pak, Ryan A. and Gray, Andrew N. and Gross, Carol A. and Dixit, Atray and Parnas, Oren and Regev, Aviv and Weissman, Jonathan S.},
  title   = {A Multiplexed Single-Cell {CRISPR} Screening Platform Enables Systematic Dissection of the Unfolded Protein Response},
  journal = {Cell},
  volume  = {167},
  number  = {7},
  pages   = {1867--1882.e21},
  year    = {2016},
  doi     = {10.1016/j.cell.2016.11.048}
}

@article{scperturbench2025,
  author  = {Wei, Zhiting and Wang, Yiheng and Gao, Yicheng and Wang, Shuguang and Li, Ping and Si, Duanmiao and Gao, Yuli and Wu, Siqi and Li, Danlu and Dong, Kejing and Yang, Xingbo and Tang, Chen and Fu, Shaliu and Chen, Xiaohan and Li, Wannian and You, Yuzhou and Zhang, Chen and Liang, Aibin and Chuai, Guohui and Liu, Qi},
  title   = {Benchmarking Algorithms for Generalizable Single-Cell Perturbation Response Prediction},
  journal = {Nature Methods},
  volume  = {23},
  number  = {2},
  pages   = {451--464},
  year    = {2026},
  doi     = {10.1038/s41592-025-02980-0}
}

@article{szklarczyk2023,
  author  = {Szklarczyk, Damian and Kirsch, Rebecca and Koutrouli, Mikaela and Nastou, Katerina and Mehryary, Farrokh and Hachilif, Radja and Gable, Annika L. and Fang, Tao and Doncheva, Nadezhda T. and Pyysalo, Sampo and Bork, Peer and Jensen, Lars J. and von Mering, Christian},
  title   = {The {STRING} Database in 2023: Protein--Protein Association Networks and Functional Enrichment Analyses for Any Sequenced Genome of Interest},
  journal = {Nucleic Acids Research},
  volume  = {51},
  number  = {D1},
  pages   = {D638--D646},
  year    = {2023},
  doi     = {10.1093/nar/gkac1000}
}
\bibliographystyle{plainnat}

\newpage\appendix

\section{Supplementary Data}
\label{sec:sup-tables}

\subsection{Datasets and Models}\label{sec:datasets}

This appendix shows the explicit details about the datasets and models used. Table~\ref{tab:datasets} summarizes the properties of each graph dataset. Table~\ref{tab:neural_config} shows the configuration of the graph neural networks baseline. Table~\ref{tab:time} compare the training cost of each experiment. Finally, we also summarizes the parameters count for each domain and dataset in Table~\ref{tab:params}.

\begin{table}[!ht]
\centering
\caption{Datasets and graphs. The model sees only the unweighted, undirected graph and the seed. Everything else is withheld. Domain~1: bus graph of in-service lines and transformers (largest connected component). A query injects $+20$\,MW\,/\,$+6$\,MVAr at the seed bus and records the AC power-flow response at every other bus (non-converged probes are dropped). gen./transf.: generator buses and transformer branches, both withheld from the model. Domain~2: largest connected component of STRING v12.0 human interactions with combined score $\geq 400$ over the measured genes (no edges added). A query is one perturbed gene (Frangieh: control arm only). In both domains all non-seed nodes are candidates, the top 10\% by response are positives, and splits are seed-disjoint five-fold partitions (per-fold query counts shown). $\rho_{\mathrm{hop}}$: Spearman correlation between $-$(hop distance to the seed) and the response, averaged over queries (angle\,/\,voltage for grids); a proximity-only baseline.}
\label{tab:datasets}
\vspace{1ex}
\small
\resizebox{\linewidth}{!}{%
\begin{tabular}{l ll rr rrrr r c r}
\toprule
Dataset & Source & Details & $|V|$ & $|E|$ & Avg.\ deg. & Density & Diam. & Queries & Train\,/\,Val\,/\,Test & $\rho_{\mathrm{hop}}$ \\
\midrule
\multicolumn{11}{l}{\textbf{Domain 1: power grids (pandapower); responses $|\Delta\theta|$ and $|\Delta V|$}} \\
case30 & IEEE 30-bus & 6 gen., 0 transf. & 30 & 41 & 2.73 & 0.0943 & 6 & 30 & 18\,/\,6\,/\,6 & 0.344\,/\,0.383 \\
case57 & IEEE 57-bus & 7 gen., 15 transf. & 57 & 78 & 2.74 & 0.0489 & 12 & 52 & 30--32\,/\,10--11\,/\,10--11 & 0.213\,/\,0.082 \\
case118 & IEEE 118-bus & 54 gen., 13 transf. & 118 & 179 & 3.03 & 0.0259 & 14 & 117 & 69--71\,/\,23--24\,/\,23--24 & 0.571\,/\,0.214 \\
case145 & IEEE 145-bus & 59 gen., 64 transf. & 145 & 422 & 5.82 & 0.0404 & 11 & 144 & 86--87\,/\,28--29\,/\,28--29 & 0.339\,/\,0.153 \\
case300 & IEEE 300-bus & 77 gen., 128 transf. & 300 & 409 & 2.73 & 0.0091 & 24 & 281 & 168--169\,/\,56--57\,/\,56--57 & 0.485\,/\,0.381 \\
case1354 & PEGASE 1354-bus & 312 gen., 151 transf. & 1{,}354 & 1{,}710 & 2.53 & 0.0019& 25 & 1{,}354 & 812--813\,/\,270--271\,/\,270--271 & 0.490\,/\,0.273 \\
\midrule
\multicolumn{11}{l}{\textbf{Domain 2: Perturb-seq on STRING PPI graphs; response $|\mathrm{logFC}|$}} \\
K562 & Replogle et al.~(2022) & CRISPRi, K562; 263k cells & 5{,}787 & 218{,}844 & 75.6 & 0.0131 & 6 & 1{,}615 & 969\,/\,323\,/\,323 & 0.125 \\
RPE1 & Replogle et al.~(2022) & CRISPRi, RPE1; 174k cells & 5{,}655 & 194{,}587 & 68.8 & 0.0122 & 6 & 1{,}329 & 797--798\,/\,265--266\,/\,265--266 & 0.098 \\
Frangieh & Frangieh et al.~(2021) & CRISPR-KO, melanoma; 27k cells & 5{,}023 & 99{,}522 & 39.6 & 0.0079 & 9 & 188 & 112--114\,/\,37--38\,/\,37--38 & 0.092 \\
Tian-i & Tian et al.~(2021) & CRISPRi, iPSC-derived neurons; 32k cells & 2{,}965 & 29{,}059 & 19.6 & 0.0066 & 10 & 178 & 106--108\,/\,35--36\,/\,35--36 & 0.127 \\
Adamson & Adamson et al.~(2016) & CRISPRi, K562; 57k cells & 4{,}950 & 105{,}599 & 42.7 & 0.0086 & 7 & 75 & 45\,/\,15\,/\,15 & 0.215 \\
\bottomrule
\end{tabular}}
\end{table}

\begin{table}[!ht]
\centering
\caption{Neural baseline configurations. Entries of the form a\,/\,b give Domain~1\,/\,Domain~2. Architecture and training protocol are shared. Depth and learning rate were selected per domain on validation Spearman $\rho$ from $\{0.005, 0.01, 0.05\} \times \{3, 6\}$ layers (Domain~1: case118, fold~0, mean over both responses; Domain~2: Adamson, fold~0) and then fixed for all datasets of that domain. The training protocol (lower block) is identical for every VQIL, RWR and neural arm, except that VQIL and RWR use learning rate $0.05$.}
\label{tab:neural_config}
\vspace{1ex}
\small
\resizebox{\linewidth}{!}{%
\begin{tabular}{l ccc}
\toprule
& GCN & GAT & Sheaf-$\mathcal{O}(2)$ \\
\midrule
\multicolumn{4}{l}{\textbf{Architecture (both domains)}} \\
Input features (50-dim) & \multicolumn{3}{c}{seed indicator, degree, log-degree, hop distance to seed (one-hot, 0--12, $\geq$13), RWSE-16, LapPE-16; one RWR-landing channel held at zero} \\
Message passing & sym.-normalised, self-loops & single-head additive attention, self-loops & $\mathcal{O}(2)$ sheaf diffusion, 2-dim stalks \\
Hidden width & 64 & 64 & 16 channels $\times$ 2 \\
Layer details & residual, LayerNorm, LeakyReLU(0.1) & residual, LayerNorm, LeakyReLU(0.1) & per-edge rotation angle from 2-layer MLP (32), learned step $\varepsilon$ \\
Dropout & 0.2 & 0.2 & 0.2 \\
Decoder & \multicolumn{3}{c}{pair head MLP$([h_v, h_s, h_v \odot h_s])$, 64 hidden, PReLU; softplus output} \\
\midrule
\multicolumn{4}{l}{\textbf{Selected per domain (Domain 1\,/\,Domain 2)}} \\
Layers & 3 / 6 & 3 / 6 & 3 / 3 \\
Learning rate & 0.01 / 0.01 & 0.005 / 0.01 & 0.005 / 0.005 \\
Trainable parameters & 28{,}546 / 41{,}410 & 28{,}930 / 42{,}178 & 15{,}019 / 15{,}019 \\
Validation $\rho$ at selection & 0.946 / 0.729 & 0.944 / 0.690 & 0.943 / 0.397 \\
\midrule
\multicolumn{4}{l}{\textbf{Training protocol (all arms, both domains)}} \\
Loss & \multicolumn{3}{c}{soft-Spearman ($\tau=1$) on log scores} \\
Optimiser & \multicolumn{3}{c}{AdamW, weight decay 0.01} \\
Batch & \multicolumn{3}{c}{8 queries (evaluation 16); at most 200 training queries sampled per epoch} \\
Epochs & \multicolumn{3}{c}{at most 300; early stopping with patience 10 on validation $\rho$; best-validation checkpoint} \\
\bottomrule
\end{tabular}}
\end{table}

\begin{table}[!ht]
\centering
\caption{Training cost ($\gamma=0.9$ for VQIL variants, $r=0.1$ for RWR). (a) Wall-clock seconds per epoch, computed as total cell time divided by epochs run (so fixed evaluation cost is included); (b) mean total seconds per cell; (c) median epochs to early stopping (at most 300, patience 10). Means and medians over the five folds (and, for Domain~1, both responses). $^\dagger$Untrained models: test-set inference time only.}
\label{tab:time}
\vspace{1ex}
\small
\resizebox{\linewidth}{!}{%
\begin{tabular}{l rrrrrr rrrrr}
\toprule
& \multicolumn{6}{c}{Domain 1: power grids} & \multicolumn{5}{c}{Domain 2: PPI graphs} \\
\cmidrule(lr){2-7}\cmidrule(lr){8-12}
Model & case30 & case57 & case118 & case145 & case300 & case1354 & K562 & RPE1 & Frangieh & Tian-i & Adamson \\
\midrule
\multicolumn{12}{l}{\textbf{(a) Seconds per epoch}} \\
Standard-VQIL & 0.20 & 0.28 & 0.62 & 0.76 & 1.49 & 3.08 & 40.1 & 34.1 & 8.49 & 3.10 & 3.57 \\
Link-phased-VQIL & 0.21 & 0.28 & 0.63 & 0.77 & 1.51 & 3.08 & 40.3 & 34.3 & 8.54 & 3.11 & 3.59 \\
Vertex-phased-VQIL & 0.21 & 0.28 & 0.63 & 0.79 & 1.54 & 3.09 & 41.1 & 34.8 & 8.68 & 3.15 & 3.65 \\
Graph-phased-VQIL & 0.20 & 0.29 & 0.64 & 0.78 & 1.56 & 3.13 & 41.2 & 35.1 & 8.72 & 3.16 & 3.66 \\
Phased-only-VQIL & 0.21 & 0.29 & 0.64 & 0.86 & 1.62 & 3.10 & 40.8 & 34.7 & 8.65 & 3.14 & 3.64 \\
\addlinespace[2pt]
Trained-RWR & 0.15 & 0.23 & 0.50 & 0.63 & 1.24 & 2.68 & 12.8 & 11.0 & 2.95 & 1.55 & 1.23 \\
\addlinespace[2pt]
GCN & 0.11 & 0.17 & 0.51 & 0.54 & 1.90 & 7.41 & 68.3 & 59.4 & 16.4 & 7.73 & 6.75 \\
GAT & 0.13 & 0.19 & 0.42 & 0.63 & 1.91 & 9.13 & 87.0 & 75.0 & 20.8 & 8.98 & 8.58 \\
Sheaf-$\mathcal{O}(2)$ & 0.13 & 0.19 & 0.63 & 0.96 & 2.83 & 14.7 & 104 & 87.8 & 21.6 & 7.27 & 9.11 \\
\midrule
\multicolumn{12}{l}{\textbf{(b) Seconds per cell (training to early stopping, including evaluation)}} \\
Naive-VQIL (untrained) & 0.02$^\dagger$ & 0.03$^\dagger$ & 0.07$^\dagger$ & 0.08$^\dagger$ & 0.17$^\dagger$ & 1.10$^\dagger$ & 12.6$^\dagger$ & 9.49$^\dagger$ & 0.82$^\dagger$ & 0.34$^\dagger$ & 0.34$^\dagger$ \\
Standard-VQIL & 8.84 & 13.8 & 21.6 & 36.0 & 54.9 & 134 & 3{,}550 & 3{,}405 & 693 & 208 & 382 \\
Link-phased-VQIL & 9.06 & 14.0 & 22.1 & 39.7 & 55.2 & 142 & 3{,}932 & 3{,}050 & 672 & 200 & 425 \\
Vertex-phased-VQIL & 8.58 & 15.0 & 32.1 & 49.2 & 84.7 & 177 & 1{,}604 & 1{,}770 & 420 & 176 & 228 \\
Graph-phased-VQIL & 9.86 & 17.0 & 42.2 & 46.5 & 70.3 & 176 & 1{,}937 & 1{,}411 & 436 & 151 & 240 \\
Phased-only-VQIL & 9.86 & 15.5 & 23.1 & 50.5 & 42.6 & 101 & 3{,}627 & 3{,}203 & 1{,}194 & 677 & 468 \\
\addlinespace[2pt]
Naive-RWR (untrained) & 0.02$^\dagger$ & 0.03$^\dagger$ & 0.07$^\dagger$ & 0.08$^\dagger$ & 0.16$^\dagger$ & 1.08$^\dagger$ & 4.87$^\dagger$ & 3.80$^\dagger$ & 0.41$^\dagger$ & 0.23$^\dagger$ & 0.17$^\dagger$ \\
Trained-RWR & 6.19 & 8.96 & 20.5 & 24.0 & 52.1 & 142 & 569 & 390 & 71.9 & 36.5 & 52.8 \\
\addlinespace[2pt]
GCN & 3.89 & 8.65 & 25.6 & 35.2 & 128 & 719 & 11{,}635 & 10{,}337 & 2{,}897 & 1{,}107 & 1{,}308 \\
GAT & 5.24 & 10.2 & 27.1 & 38.8 & 179 & 892 & 7{,}020 & 9{,}883 & 2{,}635 & 856 & 1{,}030 \\
Sheaf-$\mathcal{O}(2)$ & 5.87 & 9.98 & 44.8 & 57.9 & 284 & 1{,}284 & 9{,}451 & 6{,}849 & 2{,}013 & 922 & 861 \\
\midrule
\multicolumn{12}{l}{\textbf{(c) Median epochs to early stopping}} \\
Standard-VQIL & 42 & 44 & 34 & 46 & 35 & 42 & 94 & 108 & 74 & 72 & 108 \\
Link-phased-VQIL & 42 & 44 & 34 & 46 & 32 & 45 & 110 & 88 & 84 & 66 & 119 \\
Vertex-phased-VQIL & 41 & 44 & 49 & 62 & 54 & 52 & 37 & 50 & 44 & 57 & 67 \\
Graph-phased-VQIL & 42 & 56 & 48 & 62 & 49 & 54 & 41 & 38 & 51 & 43 & 67 \\
Phased-only-VQIL & 47 & 52 & 35 & 52 & 18 & 31 & 88 & 93 & 140 & 234 & 133 \\
\addlinespace[2pt]
Trained-RWR & 42 & 35 & 38 & 34 & 37 & 53 & 38 & 37 & 23 & 23 & 42 \\
\addlinespace[2pt]
GCN & 38 & 46 & 48 & 71 & 76 & 108 & 172 & 172 & 197 & 147 & 203 \\
GAT & 42 & 36 & 62 & 58 & 92 & 104 & 49 & 131 & 111 & 93 & 122 \\
Sheaf-$\mathcal{O}(2)$ & 44 & 49 & 66 & 59 & 101 & 88 & 98 & 80 & 95 & 119 & 95 \\
\bottomrule
\end{tabular}}
\end{table}

\begin{table}[!ht]
\centering
\caption{Trainable parameters per model and dataset ($\mathrm{K}=10^3$). VQIL and RWR learn one weight per undirected edge, plus one phase per node (vertex-phased) and/or per edge (link-phased); Phased-only-VQIL freezes the weights and trains node and edge phases. Neural baselines have a size independent of the graph but set by the depth selected per domain (Table~\ref{tab:neural_config}: 3 layers for Domain~1; 6 for GCN and GAT in Domain~2). Counts do not depend on $\gamma$ or on the response.}
\label{tab:params}
\vspace{1ex}
\small
\resizebox{\linewidth}{!}{%
\begin{tabular}{l c rrrrrr rrrrr}
\toprule
& & \multicolumn{6}{c}{Domain 1: power grids} & \multicolumn{5}{c}{Domain 2: PPI graphs} \\
\cmidrule(lr){3-8}\cmidrule(lr){9-13}
Model & Scaling & case30 & case57 & case118 & case145 & case300 & case1354 & K562 & RPE1 & Frangieh & Tian-i & Adamson \\
\midrule
\textit{Nodes} $|V|$ & & 30 & 57 & 118 & 145 & 300 & 1{,}354 & 5{,}787 & 5{,}655 & 5{,}023 & 2{,}965 & 4{,}950 \\
\textit{Edges} $|E|$ & & 41 & 78 & 179 & 422 & 409 & 1{,}710 & 218{,}844 & 194{,}587 & 99{,}522 & 29{,}059 & 105{,}599 \\
\midrule
Naive-VQIL (untrained) & -- & 0 & 0 & 0 & 0 & 0 & 0 & 0 & 0 & 0 & 0 & 0 \\
Standard-VQIL & $|E|$ & 41 & 78 & 179 & 422 & 409 & 1{,}710 & 218.8K & 194.6K & 99.5K & 29.1K & 105.6K \\
Link-phased-VQIL & $2|E|$ & 82 & 156 & 358 & 844 & 818 & 3{,}420 & 437.7K & 389.2K & 199.0K & 58.1K & 211.2K \\
Vertex-phased-VQIL & $|E|+|V|$ & 71 & 135 & 297 & 567 & 709 & 3{,}064 & 224.6K & 200.2K & 104.5K & 32.0K & 110.5K \\
Graph-phased-VQIL & $2|E|+|V|$ & 112 & 213 & 476 & 989 & 1{,}118 & 4{,}774 & 443.5K & 394.8K & 204.1K & 61.1K & 216.1K \\
Phased-only-VQIL & $|E|+|V|$ & 71 & 135 & 297 & 567 & 709 & 3{,}064 & 224.6K & 200.2K & 104.5K & 32.0K & 110.5K \\
\midrule
Naive-RWR (untrained) & -- & 0 & 0 & 0 & 0 & 0 & 0 & 0 & 0 & 0 & 0 & 0 \\
Trained-RWR & $|E|$ & 41 & 78 & 179 & 422 & 409 & 1{,}710 & 218.8K & 194.6K & 99.5K & 29.1K & 105.6K \\
\midrule
GCN & fixed & 28.5K & 28.5K & 28.5K & 28.5K & 28.5K & 28.5K & 41.4K & 41.4K & 41.4K & 41.4K & 41.4K \\
GAT & fixed & 28.9K & 28.9K & 28.9K & 28.9K & 28.9K & 28.9K & 42.2K & 42.2K & 42.2K & 42.2K & 42.2K \\
Sheaf-$\mathcal{O}(2)$ & fixed & 15.0K & 15.0K & 15.0K & 15.0K & 15.0K & 15.0K & 15.0K & 15.0K & 15.0K & 15.0K & 15.0K \\
\bottomrule
\end{tabular}}
\end{table}

\newpage
\clearpage
\subsection{Results for $\gamma = 0.7$}\label{sec:gamma}

In this appendix we show the training results using the horizon $\gamma = 0.7$. They are summarized in Table~\ref{tab:grid_g07} for the transmision grids, and in Table~\ref{tab:ppi_g07} for the PPI networks. Moreover, Table~\ref{tab:readout07} shows the readout ablation results at this horizon.

\begin{table}[!ht]
\centering
\caption{Power-grid perturbation-response ranking at the shorter readout horizon, $\gamma=0.7$ (matched restart $r=1-\gamma=0.3$ for RWR); all other settings as in Table~\ref{tab:grid}. Mean $\pm$ sd over 30 cells (six power grids $\times$ five seed-disjoint folds). The neural baselines have no horizon parameter, so their rows repeat Table~\ref{tab:grid}. \textbf{Bold}: main VQIL variant.}
\label{tab:grid_g07}
\vspace{1ex}
\small
\begin{tabular*}{\linewidth}{@{\extracolsep{\fill}} l cc cc @{}}
\toprule
& \multicolumn{2}{c}{Phase-angle response $|\Delta\theta|$}
& \multicolumn{2}{c}{Voltage response $|\Delta V|$} \\
\cmidrule(lr){2-3}\cmidrule(lr){4-5}
Model & Spearman $\rho$ & AUPRC@10\% & Spearman $\rho$ & AUPRC@10\% \\
\midrule
Naive-VQIL (untrained) & 0.441 $\pm$ 0.160 & 0.715 $\pm$ 0.104 & 0.238 $\pm$ 0.099 & 0.427 $\pm$ 0.123 \\
Standard-VQIL & 0.755 $\pm$ 0.119 & 0.735 $\pm$ 0.094 & 0.620 $\pm$ 0.099 & 0.511 $\pm$ 0.128 \\
Link-phased-VQIL & 0.755 $\pm$ 0.119 & 0.735 $\pm$ 0.094 & 0.620 $\pm$ 0.099 & 0.511 $\pm$ 0.128 \\
\textbf{Vertex-phased-VQIL} & \textbf{0.849 $\pm$ 0.085} & \textbf{0.777 $\pm$ 0.087} & \textbf{0.704 $\pm$ 0.080} & \textbf{0.548 $\pm$ 0.137} \\
Graph-phased-VQIL & 0.855 $\pm$ 0.083 & 0.780 $\pm$ 0.082 & 0.704 $\pm$ 0.076 & 0.548 $\pm$ 0.133 \\
Phased-only-VQIL & 0.769 $\pm$ 0.113 & 0.752 $\pm$ 0.100 & 0.513 $\pm$ 0.092 & 0.494 $\pm$ 0.136 \\
\midrule
Naive-RWR (untrained) & 0.436 $\pm$ 0.146 & 0.677 $\pm$ 0.097 & 0.245 $\pm$ 0.102 & 0.420 $\pm$ 0.118 \\
Trained-RWR & 0.713 $\pm$ 0.122 & 0.722 $\pm$ 0.096 & 0.581 $\pm$ 0.117 & 0.509 $\pm$ 0.139 \\
\midrule
GCN & 0.915 $\pm$ 0.071 & 0.813 $\pm$ 0.079 & 0.920 $\pm$ 0.041 & 0.789 $\pm$ 0.102 \\
GAT & 0.932 $\pm$ 0.058 & 0.830 $\pm$ 0.076 & 0.920 $\pm$ 0.041 & 0.795 $\pm$ 0.107 \\
Sheaf-$\mathcal{O}(2)$ & 0.913 $\pm$ 0.081 & 0.796 $\pm$ 0.095 & 0.908 $\pm$ 0.039 & 0.735 $\pm$ 0.116 \\
\bottomrule
\end{tabular*}
\end{table}

\begin{table}[!ht]
\centering
\caption{Gene-perturbation response ranking on protein--protein interaction graphs at the shorter readout horizon, $\gamma=0.7$ (matched restart $r=0.3$ for RWR); all other settings as in Table~\ref{tab:ppi}. Mean $\pm$ sd over 25 cells (five Perturb-seq datasets $\times$ five seed-disjoint folds). The neural baselines have no horizon parameter, so their rows repeat Table~\ref{tab:ppi}. \textbf{Bold}: main VQIL variant.}
\label{tab:ppi_g07}
\vspace{1ex}
\small
\resizebox{0.9\linewidth}{!}{%
\begin{tabular}{l cccc}
\toprule
& \multicolumn{4}{c}{Gene expression response $|\mathrm{logFC}|$} \\
\cmidrule(lr){2-5}
Model & Spearman $\rho$ & AUPRC@10\% & AUROC@10\% & NDCG@10\% \\
\midrule
Naive-VQIL (untrained) & 0.171 $\pm$ 0.057 & 0.153 $\pm$ 0.029 & 0.615 $\pm$ 0.042 & 0.323 $\pm$ 0.039 \\
Standard-VQIL & 0.554 $\pm$ 0.147 & 0.269 $\pm$ 0.062 & 0.798 $\pm$ 0.041 & 0.454 $\pm$ 0.043 \\
Link-phased-VQIL & 0.554 $\pm$ 0.147 & 0.269 $\pm$ 0.062 & 0.798 $\pm$ 0.041 & 0.454 $\pm$ 0.043 \\
\textbf{Vertex-phased-VQIL} & \textbf{0.587 $\pm$ 0.161} & \textbf{0.278 $\pm$ 0.061} & \textbf{0.819 $\pm$ 0.042} & \textbf{0.456 $\pm$ 0.041} \\
Graph-phased-VQIL & 0.587 $\pm$ 0.160 & 0.278 $\pm$ 0.060 & 0.819 $\pm$ 0.041 & 0.456 $\pm$ 0.042 \\
Phased-only-VQIL & 0.377 $\pm$ 0.101 & 0.189 $\pm$ 0.034 & 0.711 $\pm$ 0.041 & 0.358 $\pm$ 0.031 \\
\midrule
\makecell[l]{Naive-RWR (untrained)} & 0.169 $\pm$ 0.058 & 0.153 $\pm$ 0.030 & 0.613 $\pm$ 0.041 & 0.327 $\pm$ 0.037 \\
\makecell[l]{Trained-RWR} & 0.525 $\pm$ 0.140 & 0.251 $\pm$ 0.055 & 0.779 $\pm$ 0.042 & 0.441 $\pm$ 0.040 \\
\midrule
GCN & 0.582 $\pm$ 0.166 & 0.314 $\pm$ 0.077 & 0.815 $\pm$ 0.047 & 0.518 $\pm$ 0.056 \\
GAT & 0.528 $\pm$ 0.183 & 0.285 $\pm$ 0.088 & 0.782 $\pm$ 0.063 & 0.487 $\pm$ 0.070 \\
Sheaf-$\mathcal{O}(2)$ & 0.365 $\pm$ 0.141 & 0.208 $\pm$ 0.039 & 0.696 $\pm$ 0.043 & 0.411 $\pm$ 0.032 \\
\bottomrule
\end{tabular}}
\end{table}

\begin{table}[!ht]
\centering
\caption{Readout ablation at the $\gamma=0.7$ horizon: the $2\times2$ factorial over step weighting ($K = 40$) and the order in which the squared norm is taken. Same protocol, cells and layout as
Table~\ref{tab:readout}.Spearman $\rho$, mean $\pm$ sd over five seed-disjoint folds of six grids (30 cells) and five Perturb-seq datasets (25 cells). \textbf{Bold}: coherent resolvent readout of VQIL.}
\label{tab:readout07}
\vspace{1ex}
\small
\begin{tabular*}{\linewidth}{@{\extracolsep{\fill}} l cc c @{}}
\toprule
& \multicolumn{2}{c}{Power grids} & PPI networks \\
\cmidrule(lr){2-3}\cmidrule(lr){4-4}
Readout & Phase-angle $|\Delta\theta|$ & Voltage $|\Delta V|$ & Gene expression \\
\midrule
Uniform, add-then-square & 0.456 $\pm$ 0.120 & 0.299 $\pm$ 0.168 & 0.508 $\pm$ 0.144 \\
Uniform, square-then-add & 0.495 $\pm$ 0.145 & 0.310 $\pm$ 0.178 & 0.516 $\pm$ 0.141 \\
Geometric, add-then-square & \textbf{0.849} $\pm$ 0.085 & \textbf{0.704} $\pm$ 0.080 & \textbf{0.587 $\pm$ 0.161} \\
Geometric, square-then-add & 0.861 $\pm$ 0.083 & 0.715 $\pm$ 0.086 & 0.536 $\pm$ 0.153 \\
\bottomrule
\end{tabular*}
\end{table}

\newpage

\clearpage

\subsection{Learned Parameter Fields}

\begin{figure}[h]
\centering
\includegraphics[width=0.88\linewidth]{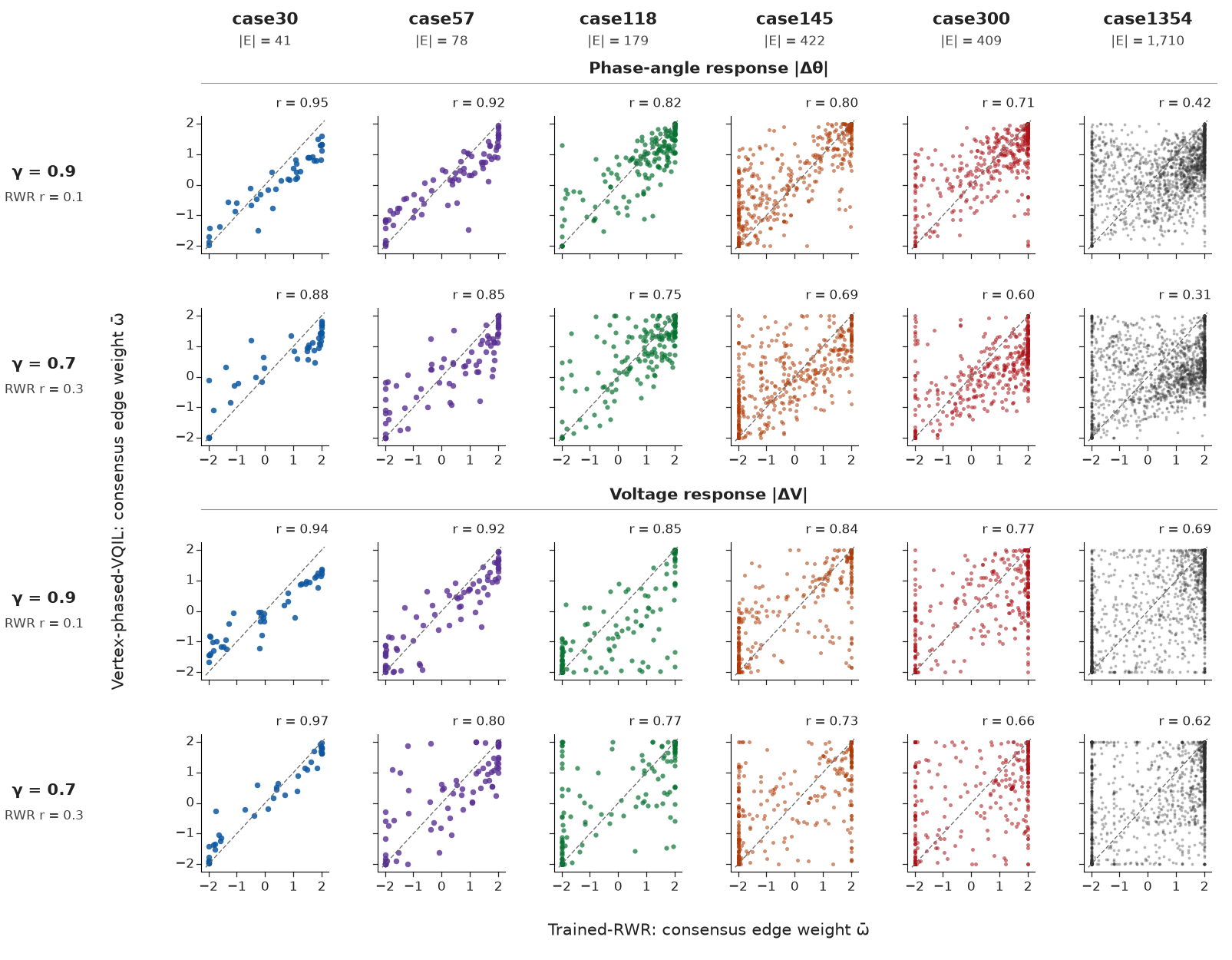}\\[4pt]
\includegraphics[width=0.88\linewidth]{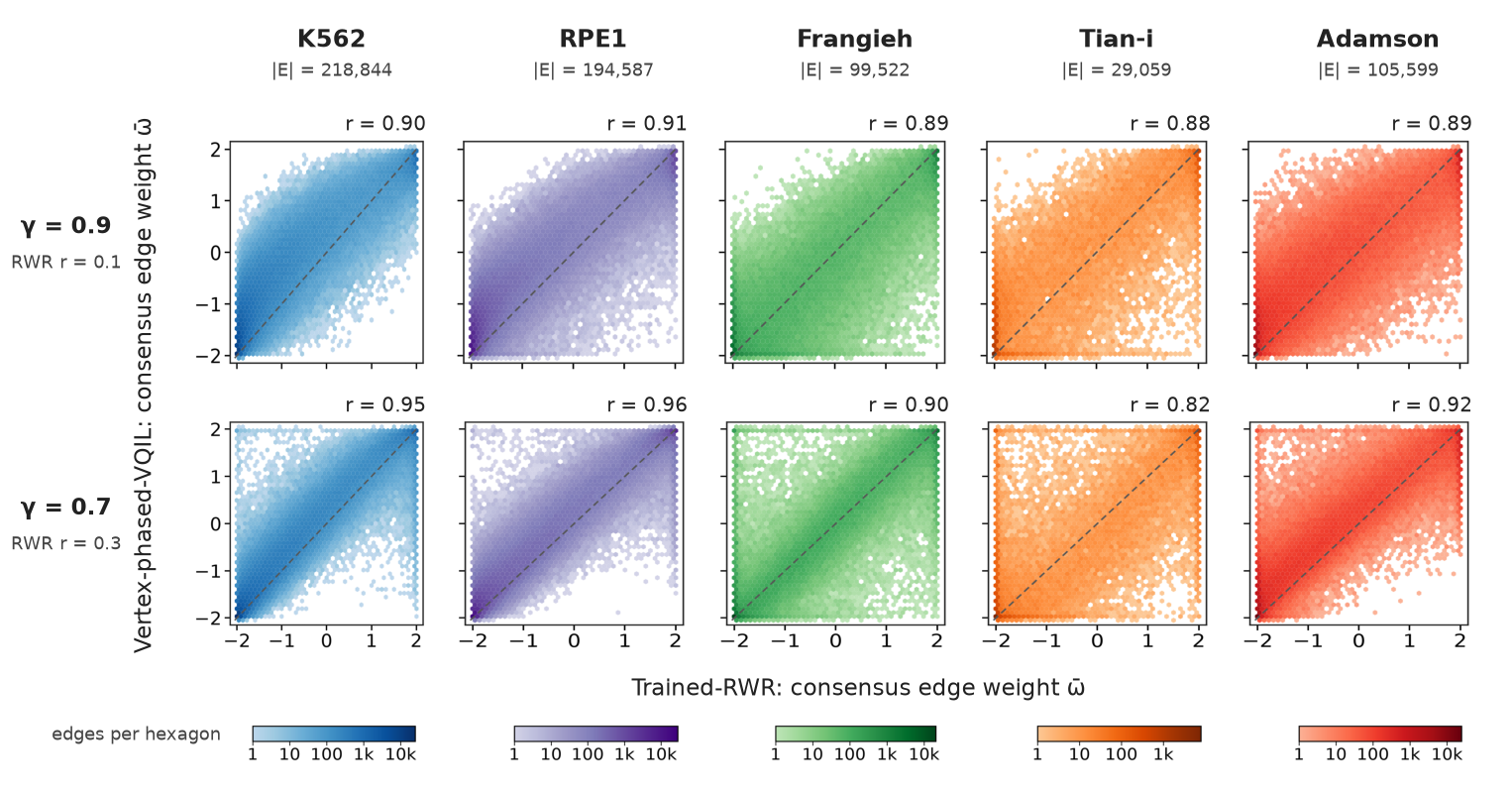}
\caption{Agreement between the edge-weight fields learned by Vertex-phased-VQIL and Trained-RWR on identical edges. Axes show consensus weights $\bar\omega$ (means over the five folds; horizontal: Trained-RWR, vertical: Vertex-phased-VQIL), the dashed line is the identity, and the annotated $r$ is the Pearson correlation between the two consensus fields. Top: power grids (one point per edge) for the phase-angle and voltage responses at $\gamma = 0.9$ (RWR restart $0.1$) and $\gamma = 0.7$ (restart $0.3$). Bottom: PPI networks at both horizons, as hexagonal bins colored by the number of edges per bin (log scale).}
\label{fig:omega_correlation}
\end{figure}

\begin{figure}[h]
\centering
\includegraphics[width=0.95\linewidth]{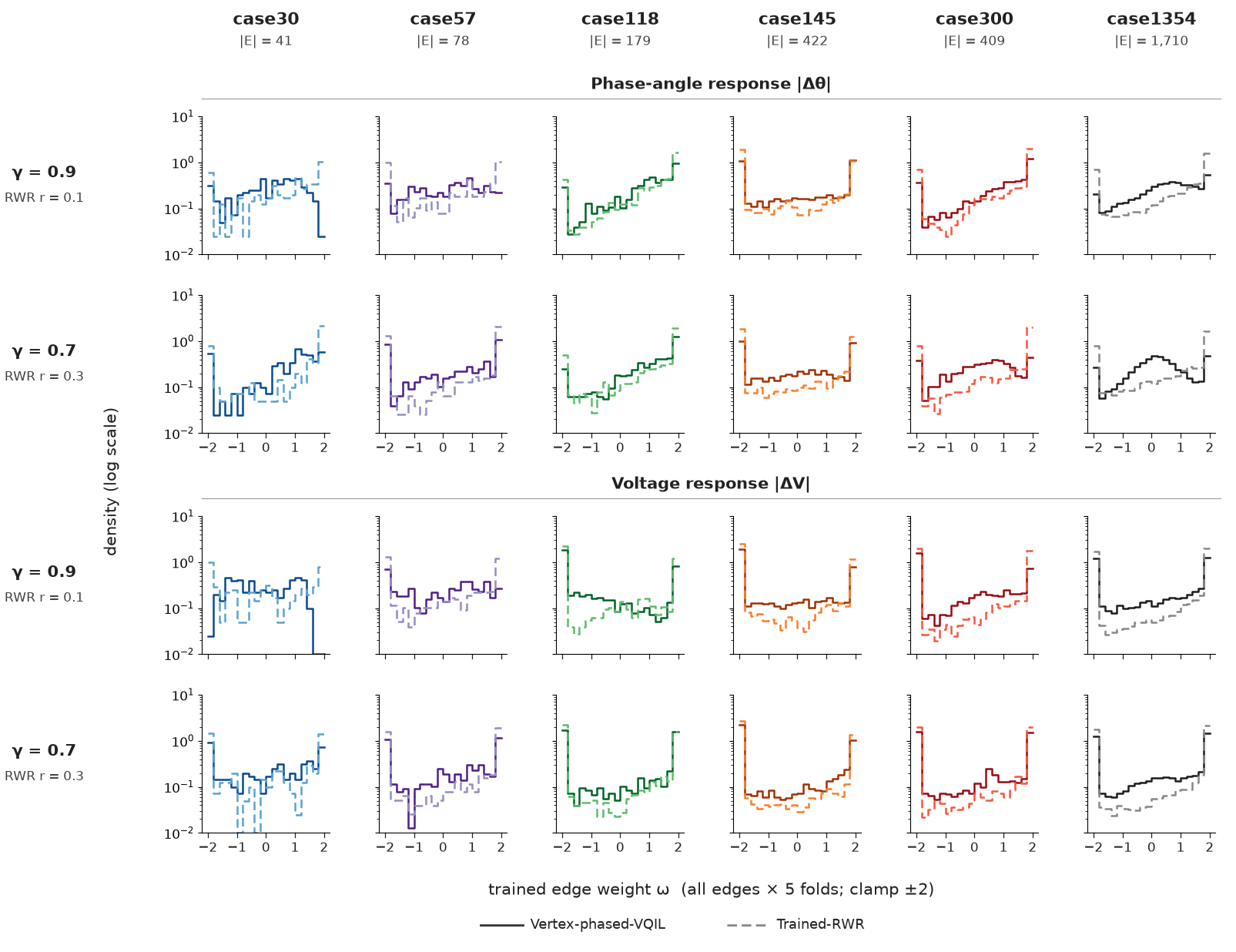}\\[4pt]
\includegraphics[width=0.95\linewidth]{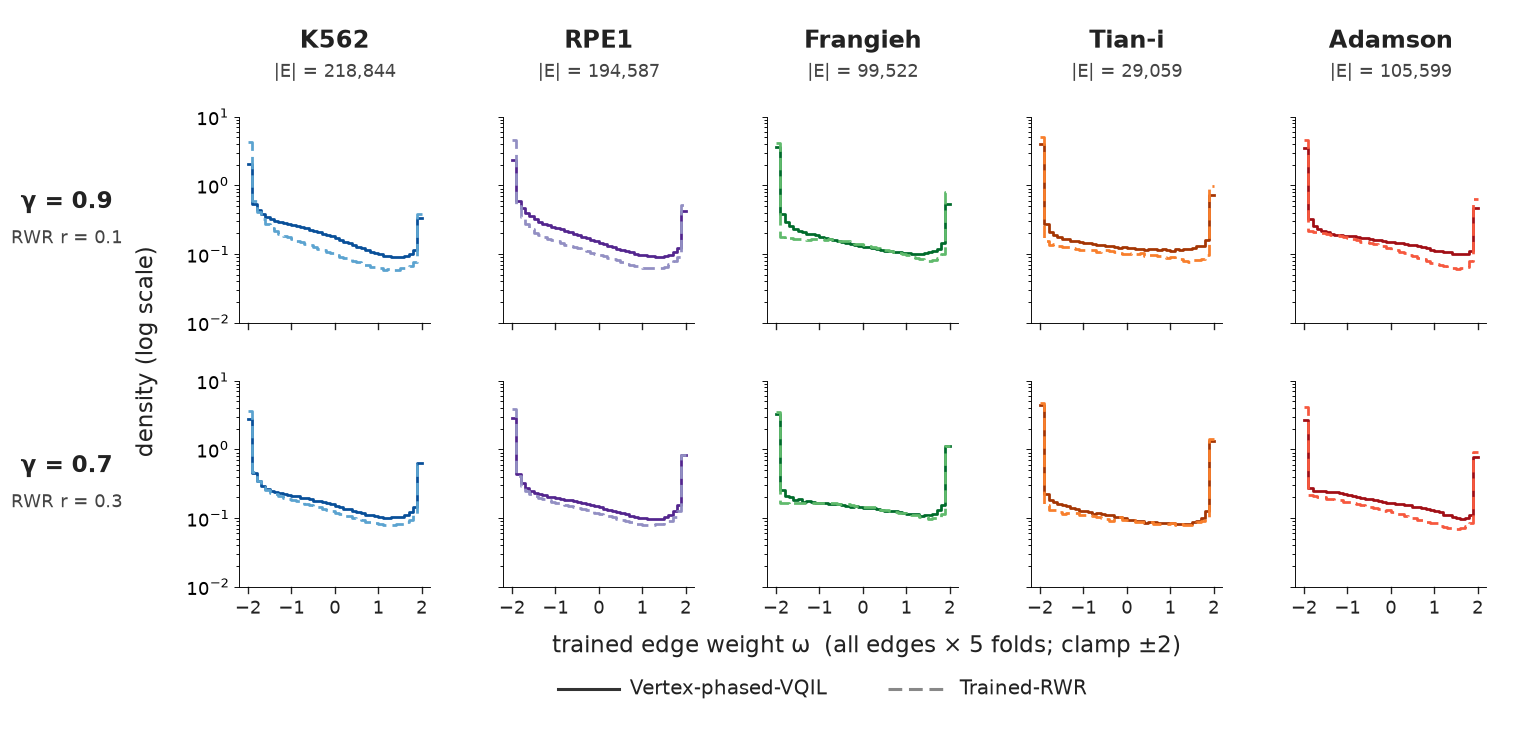}
\caption{Distributions of the trained edge weights $\omega$ of Vertex-phased-VQIL (solid) and Trained-RWR (dashed), pooled over all edges and the five folds (log-scale density; weights are clamped to $[-2, 2]$). Top: power grids (columns) for the phase-angle and voltage responses at $\gamma = 0.9$ and $\gamma = 0.7$. Bottom: PPI networks at both horizons. At $\gamma = 0.9$, Trained-RWR places $46\%$ (grids) and $43\%$ (PPI networks) of its weights at the $\pm 2$ bounds, compared with $21\%$ and $25\%$ for VQIL.}
\label{fig:omega_distribution}
\end{figure}

\begin{figure}[!ht]
\centering
\includegraphics[width=0.9\linewidth]{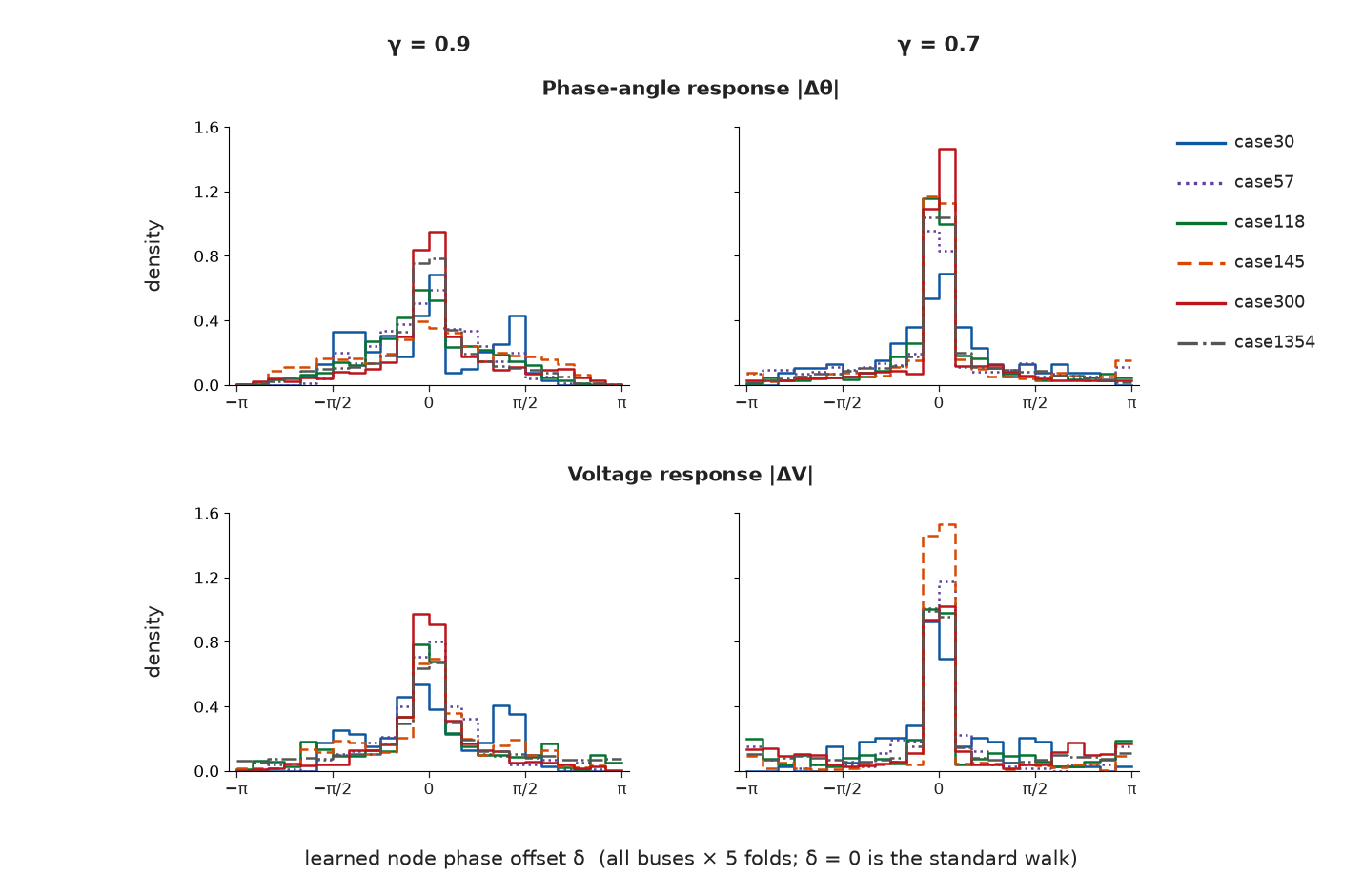}\\[4pt]
\includegraphics[width=0.9\linewidth]{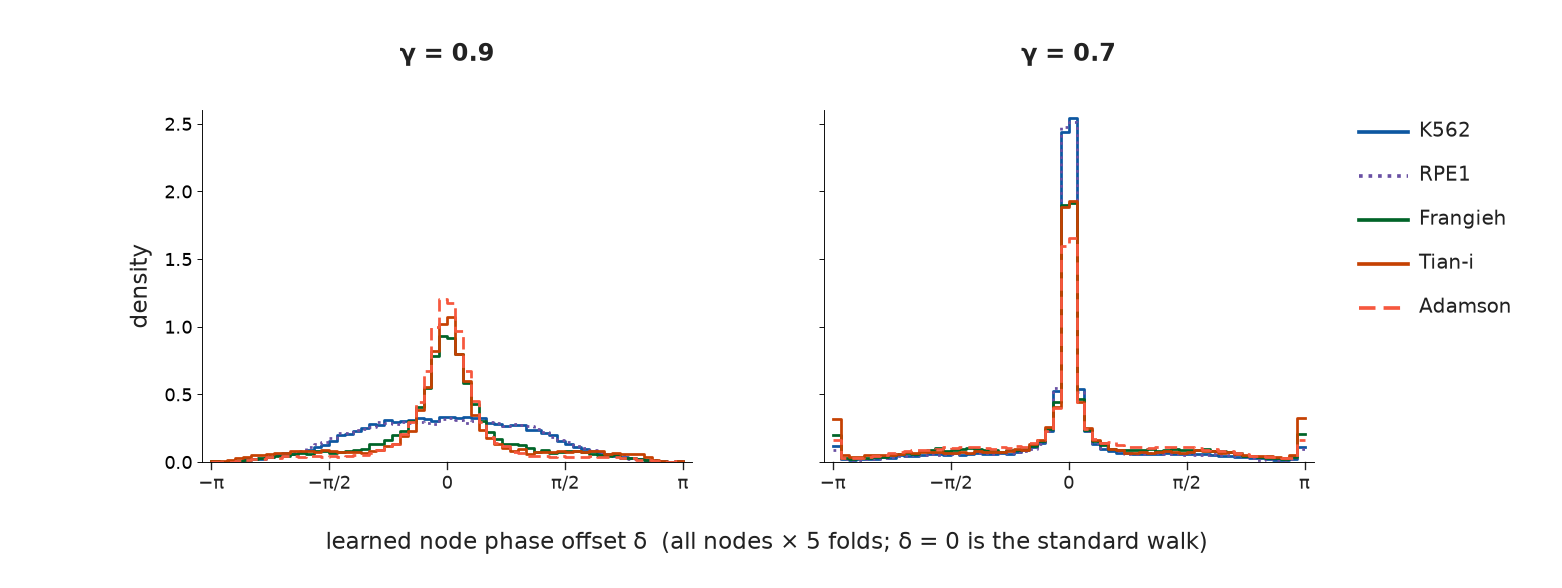}
\caption{Distributions of the learned node-phase offsets $\delta$ of Vertex-phased-VQIL (Trained-RWR has no node parameter), pooled over all nodes and the five folds, at $\gamma = 0.9$ (left) and $\gamma = 0.7$ (right); $\delta = 0$ is the standard walk. Top: power grids (one line per grid) for the phase-angle and voltage responses. Bottom: PPI networks (one line per dataset). The distributions are nearly symmetric about zero. At $\gamma = 0.9$, $0.3\%$ (grids) and $0.06\%$ (PPI networks) of the nodes lie at the $\pm\pi$ bound; at $\gamma = 0.7$, a small fraction of nodes accumulates at the bound.}
\label{fig:delta_distribution}
\end{figure}

This appendix provides additional details on the learned parameter fields discussed in Section~\ref{sec:results}. Figure~\ref{fig:omega_correlation} compares the edge weights learned by Vertex-phased-VQIL and Trained-RWR, while Figure~\ref{fig:omega_distribution} shows their corresponding weight distributions. Figure~\ref{fig:delta_distribution} shows the distributions of the learned node-phase offsets of Vertex-phased-VQIL. Finally, Figures~\ref{fig:grid-vis} and~\ref{fig:ppi-vis} visualize the trained fields together with the withheld physics and biology  for the remaining power grids and the five PPI networks, respectively.

\newcommand{\sizeone}{0.77}
\begin{figure}[!ht]
\centering
\includegraphics[width=\sizeone\linewidth]{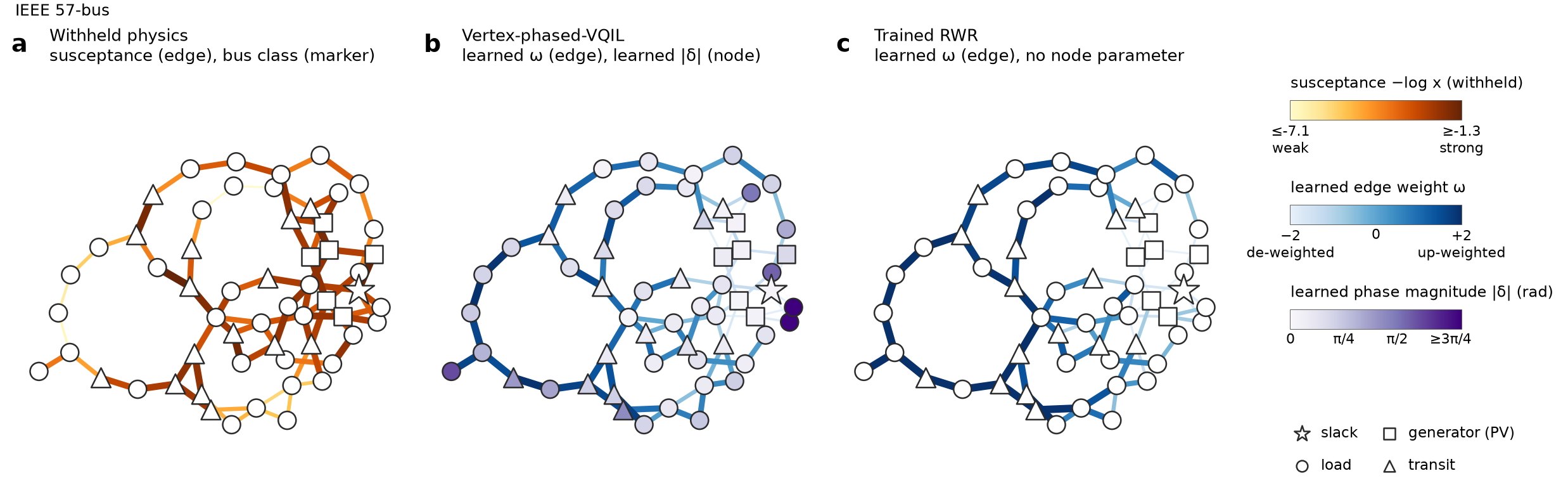}\\
\includegraphics[width=\sizeone\linewidth]{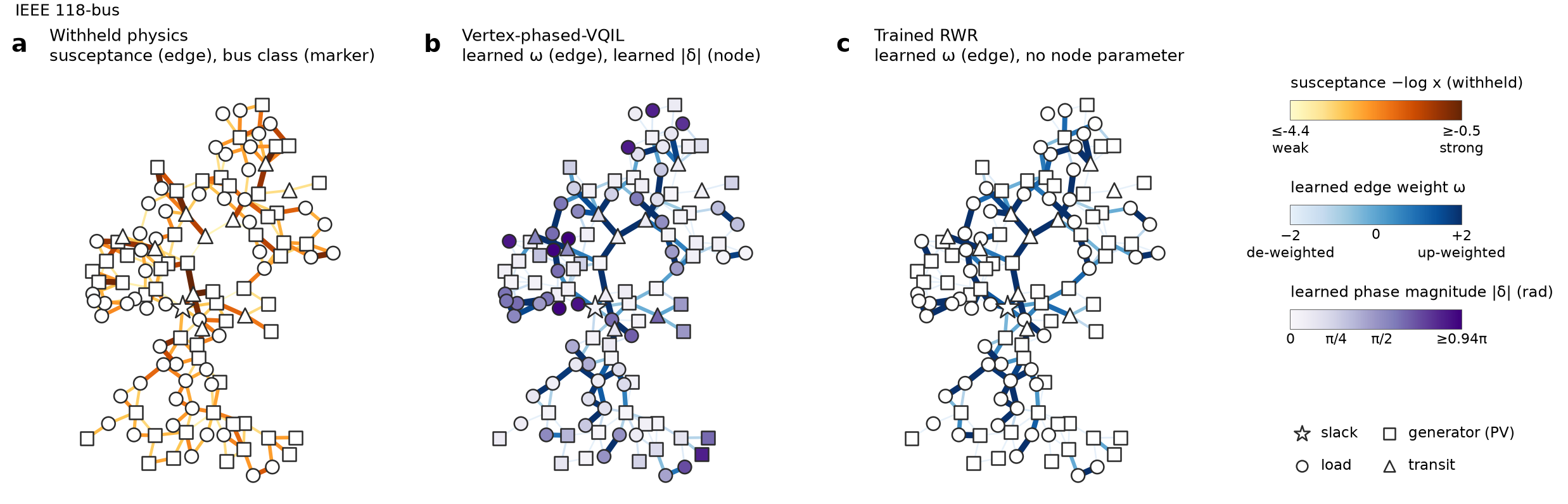}\\
\includegraphics[width=\sizeone\linewidth]{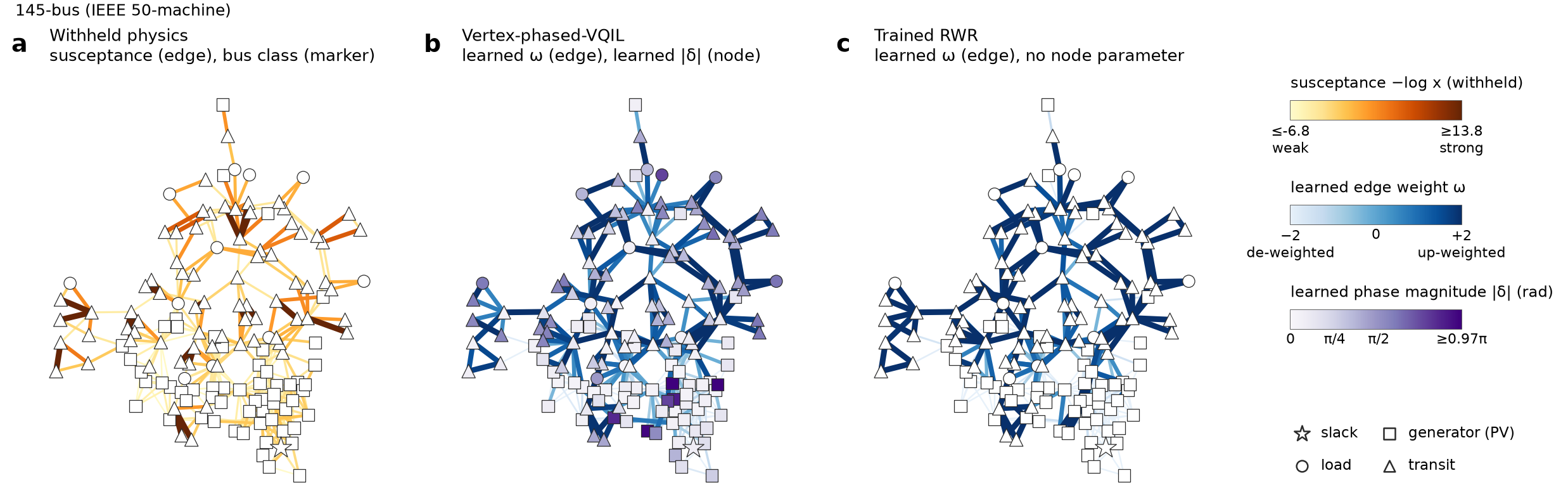}\\
\includegraphics[width=\sizeone\linewidth]{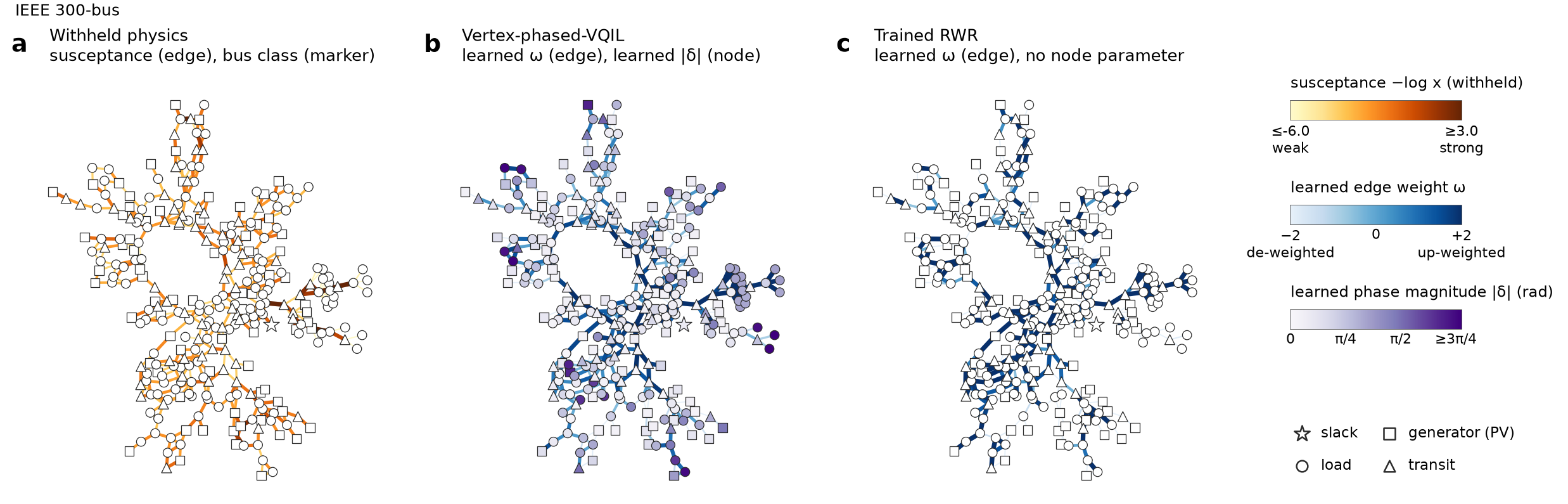}\\
\includegraphics[width=\sizeone\linewidth]{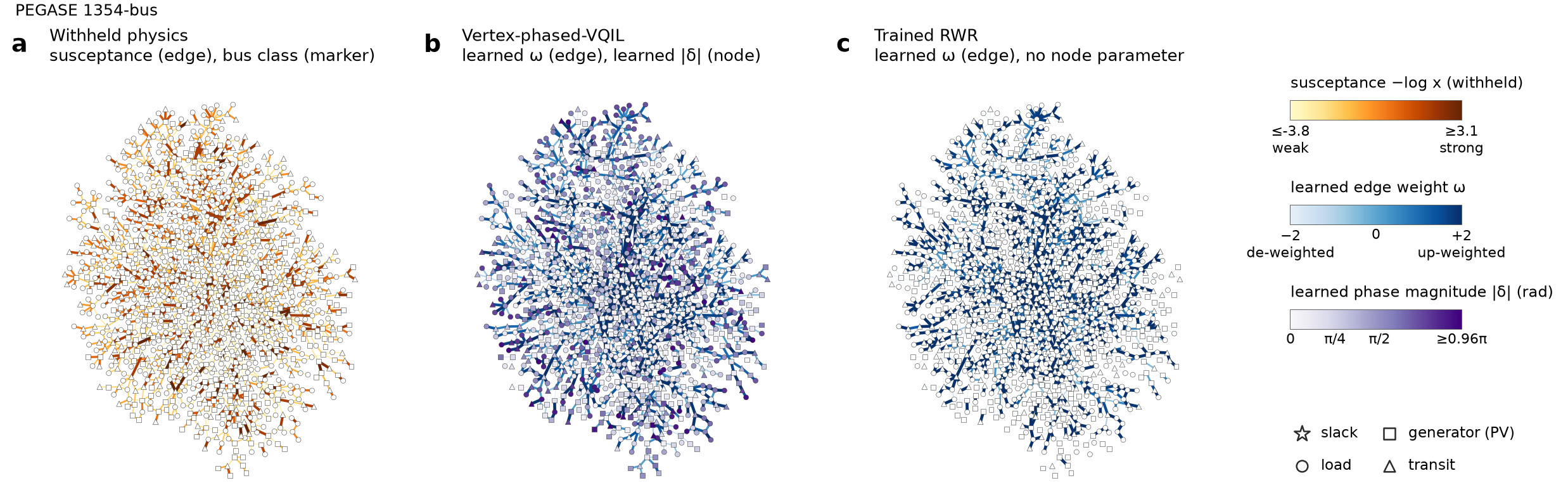}
\caption{Trained fields on the remaining five power grids (case57, case118, case145, case300, and case1354), shown as in Figure~\ref{fig:visibility} (voltage response, $\gamma = 0.9$, means over the five folds). (a) Withheld branch susceptance $-\log x$ (edge color) and bus class (marker: slack, generator, load, or transit bus). (b) Vertex-phased-VQIL: learned edge weights $\omega$ (edge color) and node-phase magnitudes $|\delta|$ (node color; the upper end of the color scale is set per grid). (c) Trained-RWR ($r = 0.1$): learned edge weights; RWR has no node parameter.}
\label{fig:grid-vis}
\end{figure}

\newcommand{\sizetwo}{0.76}
\begin{figure}[!ht]
\centering
\includegraphics[width=\sizetwo\linewidth]{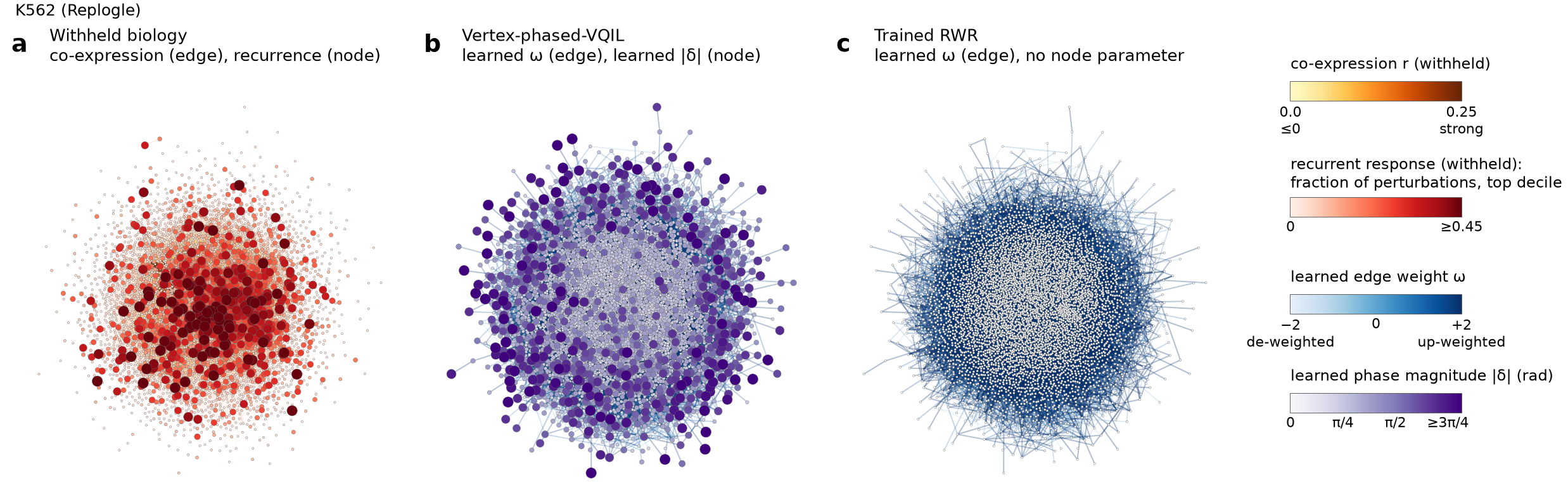}\\
\includegraphics[width=\sizetwo\linewidth]{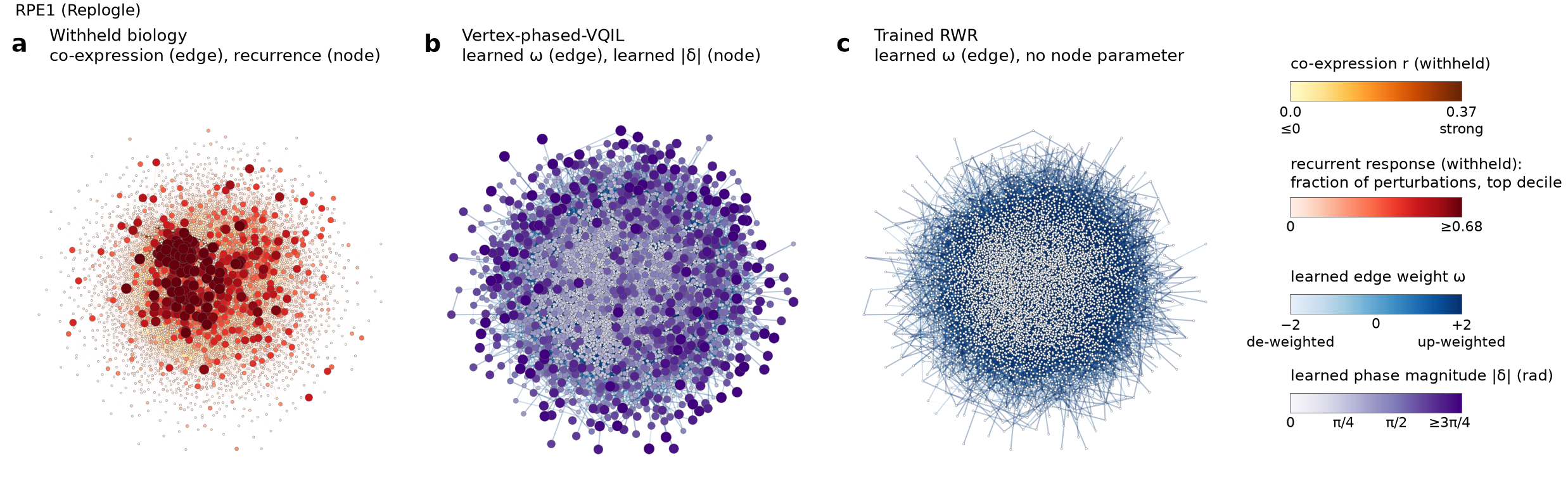}\\
\includegraphics[width=\sizetwo\linewidth]{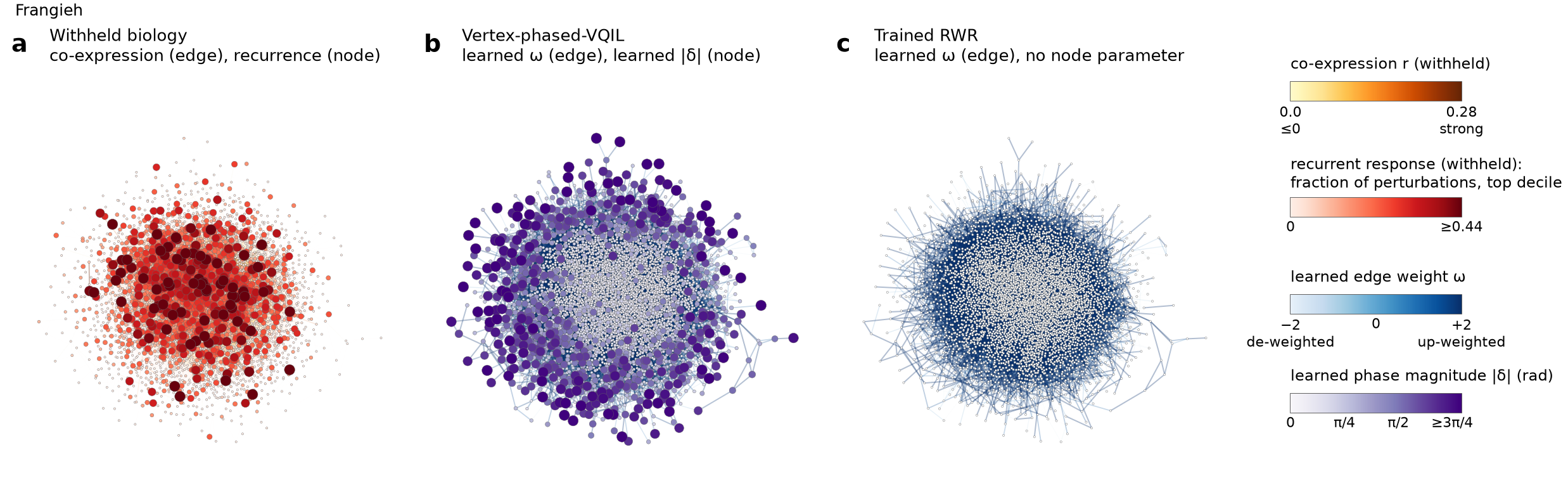}\\
\includegraphics[width=\sizetwo\linewidth]{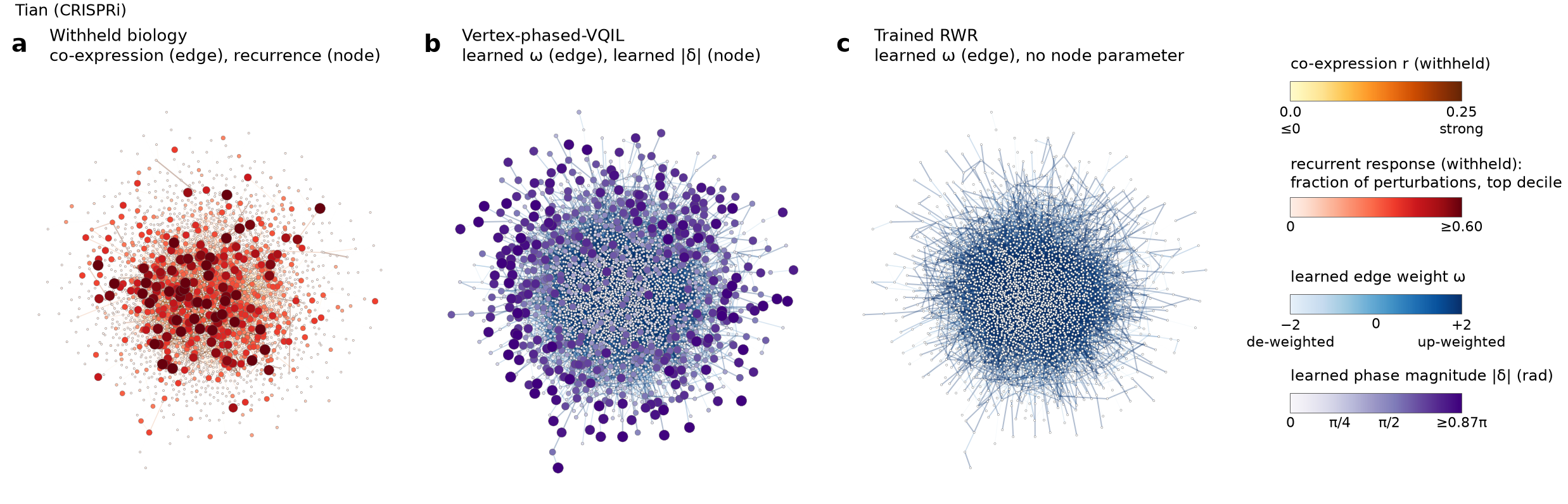}\\
\includegraphics[width=\sizetwo\linewidth]{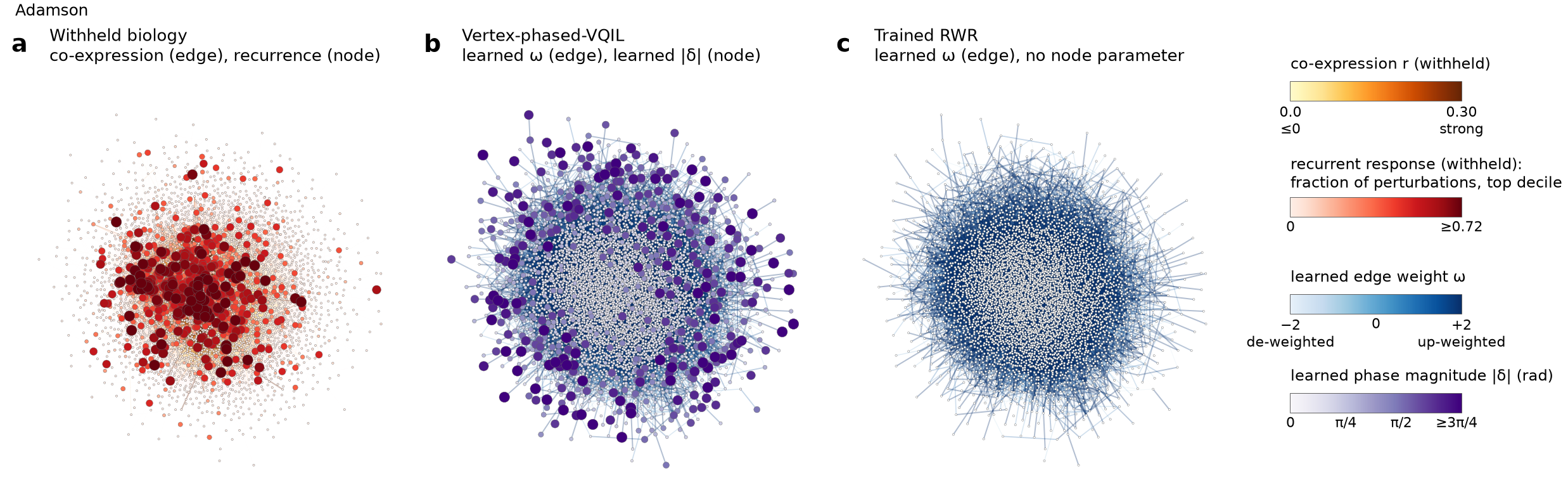}
\caption{Trained fields on the five PPI networks (K562, RPE1, Frangieh, Tian-i, and Adamson; $\gamma = 0.9$, means over the five folds). (a) Withheld biology: control-cell co-expression of the two endpoint genes (edge color) and recurrent response, the fraction of training perturbations in which a gene is among the top 10\% of responders (node color and size). (b) Vertex-phased-VQIL: learned edge weights $\omega$ (edge color) and node-phase magnitudes $|\delta|$ (node color and size). (c) Trained-RWR ($r = 0.1$): learned edge weights; RWR has no node parameter.}
\label{fig:ppi-vis}
\end{figure}

\end{document}